\documentclass[11pt]{article}

\usepackage[final]{acl}

\usepackage{times}
\usepackage{latexsym}

\usepackage[T1]{fontenc}

\usepackage[utf8]{inputenc}

\usepackage{microtype}

\usepackage{inconsolata}

\usepackage{graphicx}

\usepackage{amsmath}
\usepackage{multirow}
\usepackage{booktabs}
\usepackage{subcaption}
\usepackage{float}
\usepackage{footmisc}
\usepackage{array}
\usepackage{hyperref}
\usepackage{enumitem}
\usepackage[table]{xcolor}
\usepackage{arydshln}
\usepackage{siunitx}
\usepackage{makecell}
\title{ADVICE: Answer-Dependent Verbalized Confidence Estimation}

\author{Ki Jung Seo, Sehun Lim, Taeuk Kim\thanks{Corresponding author.} \\
Department of Computer Science, Hanyang University, Seoul, Republic of Korea \\
{\tt \{tjrlwjd1,sehun9081,kimtaeuk\}@hanyang.ac.kr}}

\begin{document}
\maketitle
\begin{abstract}
Recent progress in large language models (LLMs) has enabled them to express their confidence in natural language, improving transparency and reliability.
However, this expressiveness is often accompanied by systematic overconfidence, whose underlying causes remain poorly understood. 
In this work, we analyze the dynamics of verbalized confidence estimation and identify \textit{answer-independence}---the failure to condition confidence on the model’s own answer---as a primary driver of this behavior.
To address this, we introduce \textbf{ADVICE} (\textbf{A}nswer-\textbf{D}ependent \textbf{V}erbal\textbf{I}zed \textbf{C}onfidence \textbf{E}stimation), a fine-tuning framework that promotes answer-grounded confidence estimation.
Extensive experiments show that ADVICE substantially improves confidence calibration, while exhibiting strong generalization to unseen settings without degrading task performance.
We further demonstrate that these gains stem from enhanced answer dependence, shedding light on the origins of overconfidence and enabling trustworthy confidence verbalization.
\end{abstract}

\section{Introduction}

Recent advances in large language models (LLMs) have led to improvements in performance across diverse tasks \cite{grattafiori2024llama3herdmodels, openai2024gpt4ocard}. 
Nonetheless, hallucination---the generation of factually inaccurate or fabricated content---remains a persistent limitation \cite{Ji_2023}, with some arguing that it is theoretically unavoidable \cite{xu2024hallucination, kalai2025languagemodelshallucinate}.
This poses an obstacle to the reliable use of LLMs, particularly in high-stakes domains such as law and healthcare \cite{jayakumar-etal-2023-large, sakai2025largelanguagemodelshealthcare}.

\begin{figure}[t]
    \centering
    \includegraphics[width=\columnwidth]{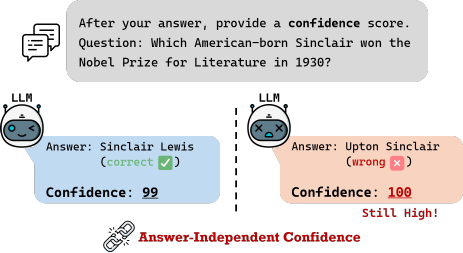}
    \caption{LLMs tend to verbalize their overconfidence \textit{irrespective of whether their answers are correct}. We propose a method (ADVICE) to mitigate this problem, achieving well-calibrated verbalized confidence.}
    \label{fig:motivation_diagram}
\end{figure}

As a remedy, recent studies refine LLMs to provide not only answers but also confidence estimates \cite{lin2022teaching, tian-etal-2023-just, xiong2024can}, aiming to manage the inherent incompleteness of LLMs rather than eliminate it entirely.
In this sense, the estimated confidence is intended to approximate the likelihood of the corresponding answer being correct \cite{pmlr-v70-guo17a}.\footnote{In related work, the terms \textbf{uncertainty} and \textbf{confidence} are often used interchangeably. For clarification, we follow the definitions of \citet{lin2024generating}: uncertainty pertains only to the input ($q$), i.e., $P(\cdot|q)$, while confidence concerns both the input and the corresponding answer ($a$), that is, $P(\cdot|q,a)$.\label{unc_conf}}
Well-calibrated models can thus express high assurance when confident and appropriately convey caution when uncertain, reinforcing their reliability.

Confidence estimation in LLMs has been explored through a range of approaches, including post-hoc extraction of confidence scores.
Among these, \textit{verbalized confidence}, which requires LLMs to articulate confidence levels in natural language during generation, has attracted sustained attention due to its universal applicability and user-friendly nature \cite{yang2025on}.
However, its broader application is hindered by the well-known issue of overconfidence \cite{xiong2024can, groot-valdenegro-toro-2024-overconfidence, leng2025taming, xu-etal-2025-language}, i.e., the tendency to assign high confidence irrespective of output quality (see Figure \ref{fig:motivation_diagram}).

In the literature, research on mitigating the overconfidence problem can be broadly categorized into three directions: prompting-based techniques, sampling-based methods such as self-consistency \cite{zhou2025steerconfsteeringllmsconfidence}, and fine-tuning \cite{li2025conftunertraininglargelanguage}.
Although such methods have contributed to improved calibration, their emphasis lies on \textit{how to mitigate overconfidence rather than why it arises}, leaving its primary causes largely unexplained.

In this work, we first investigate the intermediate process through which LLMs estimate confidence, eliciting explicit verbalization and probing their inner workings.
Specifically, we study how much the model relies on its own answer, since this property characterizes confidence and differentiates it from other measures of uncertainty (Footnote \ref{unc_conf}).\footnote{We further take inspiration from neuroscience \cite{NAVAJAS201655,10.7554/eLife.67556}, where confidence estimation is framed as post-decisional evidence accumulation.}
Our analyses reveal that LLM-generated answers and confidence verbalization seem to be internally decoupled, implying that this disjunction may underlie the poor calibration of verbalized confidence.

To further study the role of answer-groundedness in verbalized confidence estimation, we propose a novel method, \textbf{ADVICE} (\textbf{A}nswer-\textbf{D}ependent \textbf{V}erbal\textbf{I}zed \textbf{C}onfidence \textbf{E}stimation).
ADVICE explicitly encourages the model to focus more on its answer when reporting its confidence, serving as a barometer for evaluating the answer’s influence.

Through experiments, we demonstrate that ADVICE achieves performance comparable to state-of-the-art sampling-based and fine-tuning–based methods, confirming the importance of answer information in confidence estimation.
Moreover, ADVICE offers several advantages: (1) improved confidence calibration with strong generalization, (2) compact and efficient confidence representation, and (3) no compromise in task performance.

Lastly, we revisit our initial internal analysis with ADVICE-enhanced LLMs and show that performance gains are causally driven by stronger answer dependence, supporting our central claim.

\section{Related Work}

\paragraph{Verbalized confidence}
Since \citet{lin2022teaching} introduced verbalized confidence estimation, numerous studies have explored its potential, highlighting its model-agnostic design, cost-effectiveness, and accessibility to model knowledge \cite{yang2025on}.
In particular, a broad spectrum of work has sought to improve its calibration. 
As an initial direction, post-hoc methods that do not require model modification---such as prompting-based and sampling-based ones \cite{zhao-etal-2024-fact,yang2025on,zhou2025steerconfsteeringllmsconfidence}---have been proposed.
In contrast, several studies \cite{tian-etal-2023-just, DBLP:journals/corr/abs-2503-02623, li2025conftunertraininglargelanguage} adopt fine-tuning methods, specifically for the task of question answering (QA).
However, prior studies have mainly centered on developing new methods for achieving quantitative improvements, with limited qualitative analysis of the underlying mechanisms.
To fill this gap, we present an in-depth investigation into the operational mechanism of verbalized confidence estimation and introduce an intuitive method.

\paragraph{LLM probing methods}

With the wide adoption of LLMs, understanding their inner workings has become crucial, leading to a surge of research on their mechanistic interpretability \cite{mohammadi2025explainabilitypracticesurveyexplainable}.
In particular, a line of work on controlling and analyzing the attention mechanism, e.g., Attention Rollout, Attention Flow, and Attention Knockout \cite{abnar-zuidema-2020-quantifying,geva2023dissecting}, has gained interest.
Meanwhile, gradient-based attribution methods provide a more direct quantification of output sensitivity to input perturbations. 
Integrated Gradients \cite{Sundararajan2017AxiomaticAF} attributes output importance to input tokens by integrating gradients along the path from a baseline to the input.
We employ methods from both paradigms to probe the relationship between verbalized confidence estimation and the model’s answer.

\section{Claim: Verbalized Confidence is Nearly Answer-Independent}
\label{sec:Claim: Verbalized Confidence is Nearly Answer-Independent}

By definition, verbalized confidence should reflect a model’s degree of belief in its generated answer.
To examine whether this causal relationship holds in practice, we perform two evaluations: (1) a comparison of confidence distributions conditioned on alternative answer candidates, and (2) an attribution-based analysis.
The results reveal that counterintuitively, verbalized confidence is nearly independent of whether the answer is correct.

\subsection{Comparison of Confidence Distributions}
\label{sec:comparison_distributions}
Let $q\in Q$ represent a question, and $A_q$ indicates the set of all possible answer predictions for the given question, including both factually correct and incorrect ones.
$C$ denotes the set of confidence expressions, such as \texttt{0} (very low) to \texttt{9} (very high).\footnote{We assume that the model expresses confidence in a discrete form using numerical or verbal tokens (refer to \S\ref{sec:settings}).} 

We investigate whether verbalized confidence is independent of the answer by testing the following:
\begin{align*}
\label{eq1}
P_M(C \mid q, a_i) \approx P_M(C \mid q,a_j) \;\; \forall a_i \neq a_j \in A_q,
\end{align*}
where $P_M{(C \mid \cdot)}$ represents the probability distribution over confidence expressions computed by the model $M$.\footnote{For computational efficiency, we restrict $A_q$ to 30 answers generated by $M$ using top-$p$ sampling.}
If the above equation holds, the Jensen–Shannon divergence (JSD) \cite{MENENDEZ1997307} between the left-hand side (LHS) and the right-hand side (RHS) should approach zero:
\begin{align*}
\mathrm{JSD}(P_M(C \mid q, a_i) \mid\mid P_M(C \mid q,a_j)) \approx 0.
\end{align*}
To characterize trends in JSD values across combinations of $q$, $a_i$, and $a_j$, we compute $\sum_{q \in Q}{\binom{|A_q|}{2}}$ JSD scores for three datasets (TriviaQA, MMLU, LogiQA) and visualize their distributions.
We apply this process to \textsc{Gemma-2-9b-it} and \textsc{Llama-3.1-8B-Instruct} (see Appendix~\ref{appendix:experimental_setting} for details).

\begin{figure}[t]
\centering
\begin{subfigure}{0.5\columnwidth}
\includegraphics[width=\columnwidth]{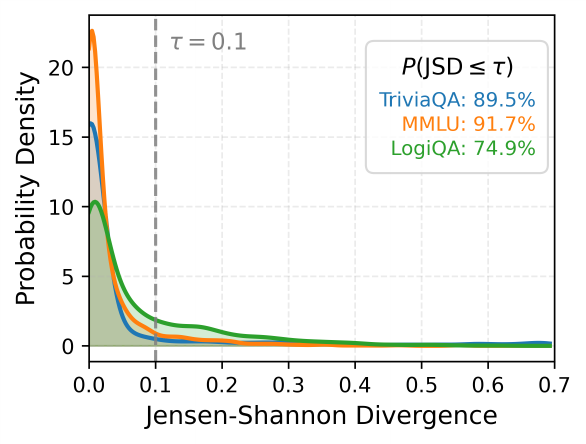}
\caption{\textsc{Gemma2-9b-it}}
\end{subfigure}\hfill
\begin{subfigure}{0.5\columnwidth}
\includegraphics[width=\columnwidth]{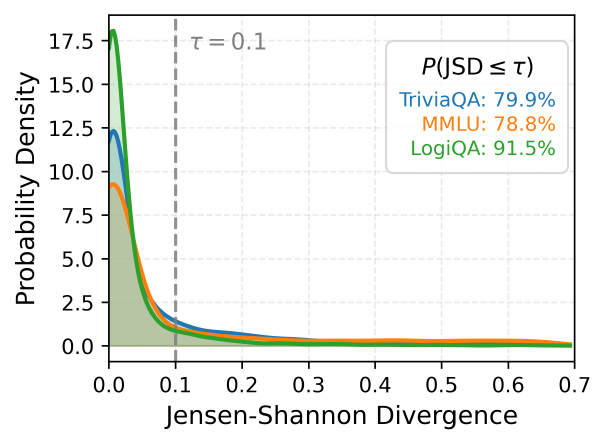}
\caption{\textsc{Llama3.1-8B-Instruct}}
\end{subfigure}
\caption{
Probability density functions (PDFs) of the set $\{\mathrm{JSD}(P_M(C | q, a_i) || P_M(C | q,a_j))\}$, where $M \in \{\textsc{Gemma2}, \textsc{Llama3.1}\}$ and $(q,a_i,a_j)$ are from TriviaQA, MMLU, LogiQA.
Each PDF (solid curve) is computed via Gaussian kernel density estimation.
Near-zero concentration implies answer-independent confidence.
}
\label{fig:answer_independence}
\end{figure}

In Figure \ref{fig:answer_independence}, we observe (1) strong concentrations of JSD scores near zero with long right tails and (2) high densities of samples in the region where JSD $\leq 0.1$.\footnote{We set the threshold $\tau = 0.1$ to compute the fraction of samples with near-zero divergence, following prior work \cite{asi8020028, 10.1007/s10994-024-06698-6} that defines two distributions as similar when their JSD is $\leq 0.1$.}
Overall, these results suggest that confidence estimates vary minimally across different answers, indicating limited use of answer-specific information. 
Results under extra experimental settings are reported in Figures \ref{fig:answer_independence_scorenumber}, \ref{fig:answer_independence_scoreletter_correct_wrong}, and \ref{fig:correct_wrong_confidence_estimation}.

\subsection{Attribution-Based Analysis}
\label{sec:attribution}
While the findings in \S\ref{sec:comparison_distributions} are striking, they warrant further validation through additional evidence from alternative analytical perspectives. 
To this end, we employ two attribution methods: Attention Rollout and Integrated Gradients.

\begin{figure}[t!]
\centering
\begin{subfigure}{0.5\columnwidth}
\includegraphics[width=\columnwidth]{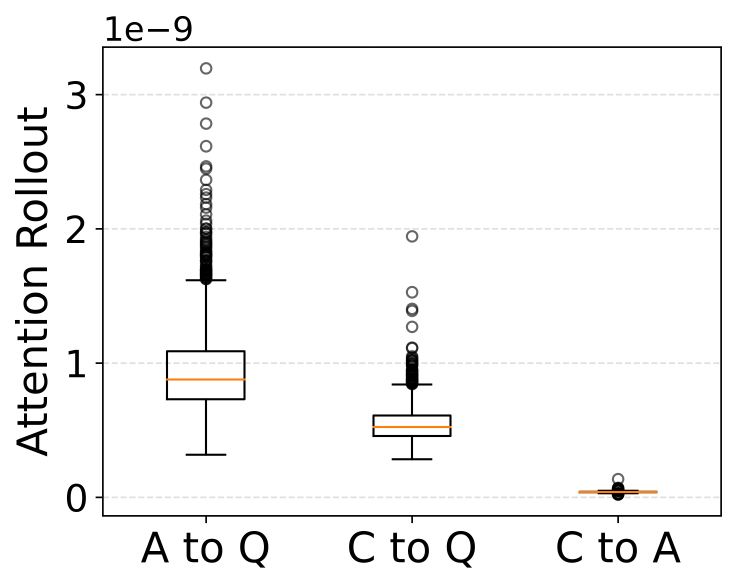}
\caption{\textsc{Gemma2-9b-it}}
\end{subfigure}\hfill
\begin{subfigure}{0.5\columnwidth}
\includegraphics[width=\columnwidth]{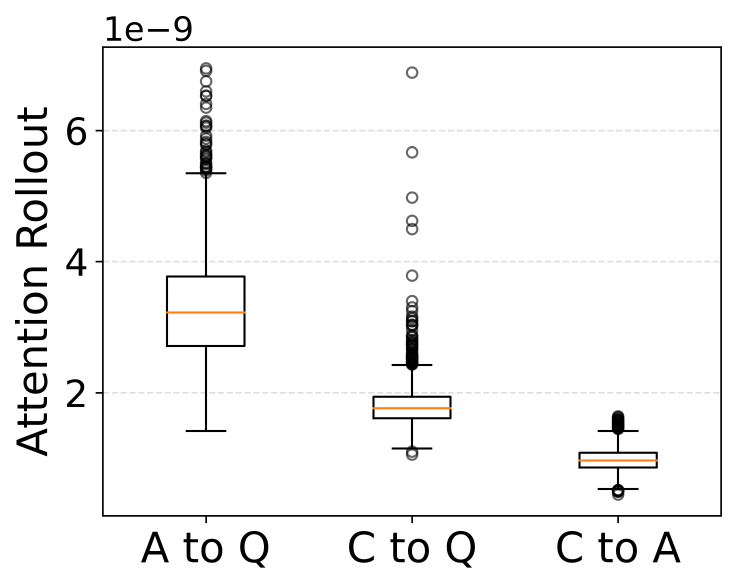}
\caption{\textsc{Llama3.1-8B-Instruct}}
\end{subfigure}
\caption{Comparison of Attention Rollout scores on three attention directions: (1) Answer to Question, (2) Confidence to Question, and (3) Confidence to Answer.}
\label{fig:gemma2_a2q_c2q_c2a}
\end{figure}

\paragraph{Attention Rollout}
Attention Rollout (AR) \cite{abnar-zuidema-2020-quantifying} quantifies the contribution of input tokens to model predictions by recursively aggregating attention weights across layers.\footnote{We refer readers to Appendix \ref{sec:appendix_probing_methods} for algorithmic details.}
We use AR to analyze how different components of the input prompt---the question ($Q$), answer ($A$), and verbalized confidence ($C$)---interact through attention inside the model.
Specifically, we examine attention from $C$ to $A$ ($C \rightarrow A$) and compare its average AR score against other attention flows, such as $A \rightarrow Q$ and $C \rightarrow Q$.
As shown in Figure \ref{fig:gemma2_a2q_c2q_c2a}, the AR score of $C \rightarrow A$ is significantly lower than those of the reference cases, suggesting that LLMs rely less on answer-specific information when estimating confidence.

\begin{figure}[t]
\centering
\includegraphics[width=\columnwidth]{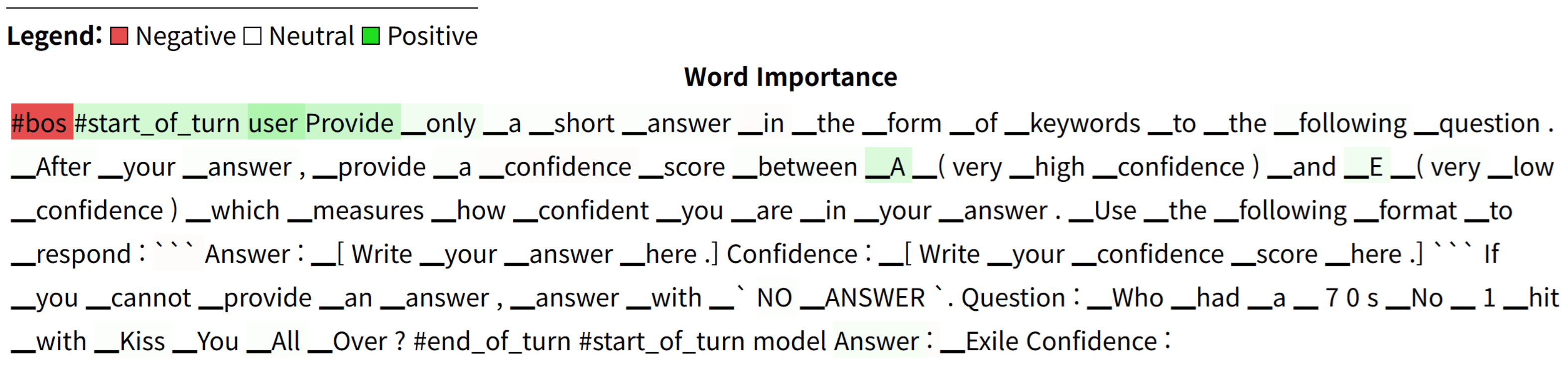}
\caption{Visualization of token attribution with Integrated Gradients (\textsc{Gemma2-9b-it}).}
\label{fig:integrated_gradients_default}
\end{figure}

\begin{figure*}[t]
    \centering
    \includegraphics[width=\textwidth]{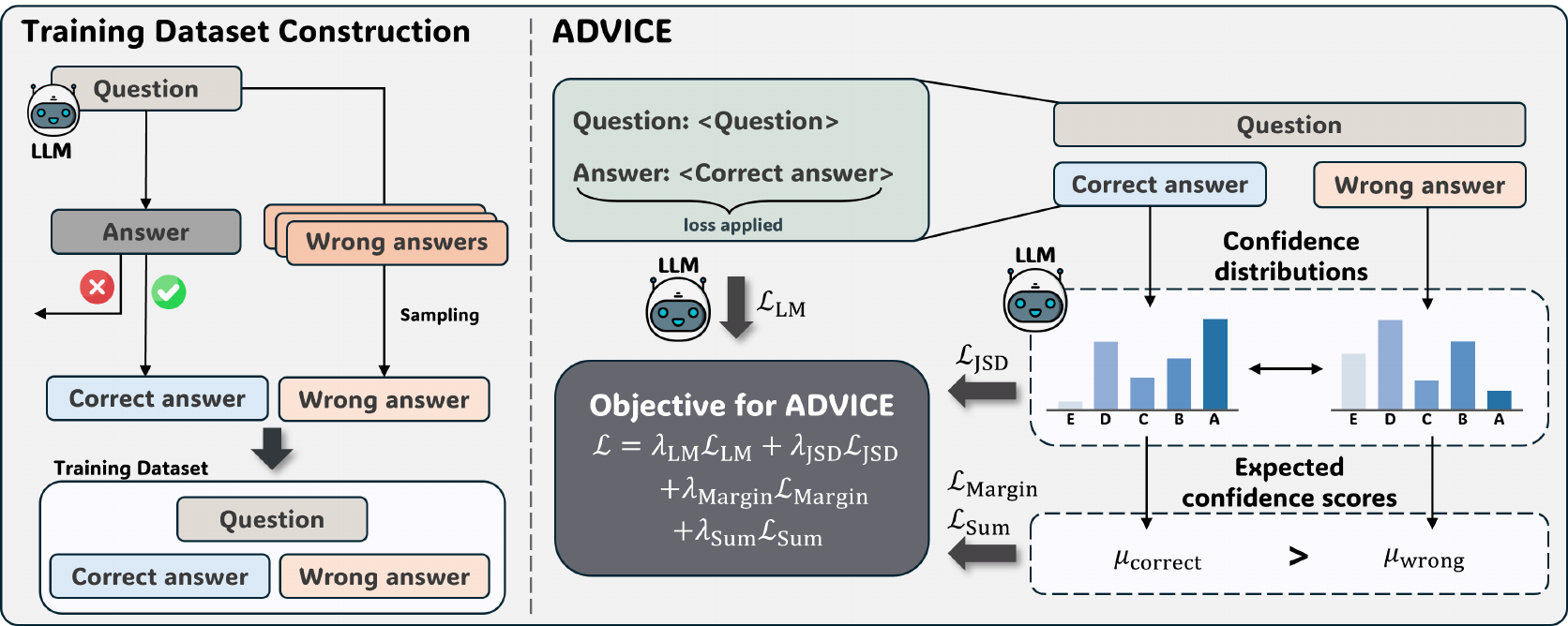}
    \caption{Illustration of the proposed \textbf{ADVICE} (\textbf{A}nswer-
\textbf{D}ependent \textbf{V}erbal\textbf{I}zed \textbf{C}onfidence \textbf{E}stimation) framework.}
    \label{fig:pipeline_diagram}
\end{figure*}

\paragraph{Integrated Gradients}
While Attention Rollout captures attention-level interactions, Integrated Gradients (IG) provides a gradient-based perspective that enables a qualitative analysis of how different input components contribute to verbalized confidence.
Figure \ref{fig:integrated_gradients_default} presents the attribution scores assigned to individual input tokens.
We observe that answer tokens are consistently under-weighted compared to tokens in other components, such as ``\texttt{user}'' and the BOS token.

\paragraph{Takeaways} 
Our empirical analyses confirm that the generation of verbalized confidence operates substantially independently of cues from the answer component, contrary to its intended definition.
As a result, we argue that this phenomenon constitutes a primary factor underlying poor calibration and overconfidence in verbalized confidence.

\section{ADVICE: Answer-Dependent Verbalized Confidence Estimation}

We present \textbf{ADVICE} (see Figure \ref{fig:pipeline_diagram}), a lightweight training framework designed to reinforce answer-groundedness in verbalized confidence estimation.

\subsection{Training Dataset Construction}
\label{sec:training dataset}

\begin{table*}[t]
\scriptsize
\centering
\setlength{\tabcolsep}{2.7pt}
\begin{tabular}{llll}
\toprule
\textbf{Question} & \textbf{Correct answer ($a_\text{correct}$)} & \textbf{Wrong answer candidates} & \textbf{Wrong answer ($a_\text{wrong}$)} \\
\midrule
\makecell[l]{Which state renewed Mike Tyson's \\boxing license in 1998?} & Nevada & [Connecticut, Oregon, California] & California \\ 
\midrule
\makecell[l]{Who were the first team to field an all \\ foreign starting line up in the English Premiership?} & Chelsea & [West Ham United, Aston Villa] & West Ham United \\
\midrule
On a standard keyboard, which is the largest key? & Space bar & [Shift key, Enter key] & Shift key \\
\midrule
\makecell[l]{Which London railway station has the most \\ platforms?} & London Waterloo & \makecell[l]{[London Victoria, London Paddington,\\ London Liverpool]} & London Victoria \\
\midrule
\makecell[l]{If a month has a Friday the thirteenth then on what \\day of the week would that month begin?} & Sunday & [Wednesday, Thursday, Monday, Friday, Tuesday] & Thursday \\
\midrule
\makecell[l]{Which major British newspaper closed down \\
for almost a year in 1978?} & The Times & [Daily Mirror, The Sun, News of the World] & The Sun \\
\bottomrule
\end{tabular}
\caption{Examples of triplets $(q,a_{\text{correct}},a_{\text{wrong}})$ in our training dataset.
The diversity of wrong answers encourages the model to learn fine-grained distinctions between correct information and subtly incorrect alternatives.}
\label{tab:example_training_dataset}
\end{table*}

We adopt TriviaQA \cite{joshi-etal-2017-triviaqa} as our training dataset, which is an open-domain, free-form question answering benchmark.
We begin by sampling 4,000 instances from the training split of the dataset.
Subsequently, we retain only instances where the model generates the correct answer under greedy decoding.
To encourage the model to express high confidence for correct answers and low confidence for incorrect ones, we construct, for each question $q$, a pair consisting of a correct answer ($a_{\text{correct}}$) and an incorrect answer ($a_{\text{wrong}}$), yielding a triplet $(q,a_{\text{correct}},a_{\text{wrong}})$.
The incorrect answer ($a_{\text{wrong}}$) is randomly sampled from the model’s responses using stochastic decoding.
Notably, stochastic decoding frequently yields hard negatives---answers that are semantically plausible and contextually relevant to the question but factually incorrect---encouraging the model to learn fine-grained distinctions.
Finally, as verbalized confidence can appear in various formats as described in \S\ref{sec:settings}, we construct two variants for each instance to train a model capable of fluently expressing confidence in multiple forms.
Examples of the triplets $(q,a_{\text{correct}},a_{\text{wrong}})$ are provided in Table \ref{tab:example_training_dataset}.

\subsection{Training Objectives}
\label{sec:training_objective}
Motivated by our findings in \S\ref{sec:Claim: Verbalized Confidence is Nearly Answer-Independent}, we train the model to explicitly condition its confidence on the generated answer while preserving its performance on general tasks.
Specifically, we define four training objectives for each triplet obtained from \S\ref{sec:training dataset}:
\begin{equation*}
\label{eq4}
\mathcal{L}_{\mathrm{LM}}= \frac{1}{|a_\text{correct}|} \sum_{x_t \in a_\text{correct}}{-\log{P(x_t \mid x_{<t}})},
\end{equation*}
\begin{equation*}
\label{eq5}
\mathcal{L}_{\mathrm{JSD}} = \max(0, \delta_{\mathrm{JSD}} - D_{\mathrm{JSD}}(P_{\text{correct}} \mid\mid P_{\text{wrong}})),
\end{equation*}
\begin{equation*}
\label{eq6}
\mathcal{L}_{\mathrm{Margin}} = \max(0,\delta_{\mathrm{Margin}}-(\mu_\text{correct} - \mu_{\text{wrong}})),
\end{equation*}
\begin{equation*}
\label{eq7}
\mathcal{L}_{\mathrm{Sum}} = |1- (\mu_\text{correct} + \mu_{\text{wrong}})|,
\end{equation*}
where $\mathcal{L}_{\mathrm{LM}}$ denotes the negative log-likelihood of the correct answer $a_\text{correct}$, added to preserve general task (e.g., QA) abilities as in \citet{li2025conftunertraininglargelanguage}.

$\mathcal{L}_{\mathrm{JSD}}$ explicitly drives the model to learn contrasting confidence distributions ($P_{\text{correct}}$ and $P_{\text{wrong}}$) for the correct ($a_{\text{correct}}$) and wrong answers ($a_\text{wrong}$) given the same question $q$.
Here, $P_\text{correct}$ and $P_\text{wrong}$ denote $P_M(C \mid q,a_\text{correct})$ and $P_M(C \mid q,a_\text{wrong})$, respectively.
However, $\mathcal{L}_{\mathrm{JSD}}$ provides no directional constraint, implying that it may still converge even if the model erroneously assigns greater confidence to incorrect answers while underestimating correct ones.
To resolve this, we apply $\mathcal{L}_{\mathrm{Margin}}$, formulated as the difference between the expected confidence assigned to correct answers ($\mu_{\text{correct}}$) and that assigned to incorrect ones ($\mu_{\text{wrong}}$).
Additionally, we introduce $\mathcal{L}_{\mathrm{Sum}}$ to enforce the ideal constraint $\mu_{\text{correct}} + \mu_{\text{wrong}} = 1$, reflecting that confidence should represent the likelihood of the answer being correct \cite{pmlr-v70-guo17a}.
Hyperparameters $\delta_{\mathrm{JSD}}$ and $\delta_{\mathrm{Margin}}$ are set to control the extent to which the model differentiates between correct and incorrect answers.\footnote{Refer to Appendix \ref{appendix:implementation_details} for training details.}

Finally, we define the total training objective:
\begin{equation*}
\label{eq8}
\begin{aligned}
\mathcal{L}
  &= \lambda_{\mathrm{LM}}\mathcal{L}_{\mathrm{LM}}
   + \lambda_{\mathrm{JSD}}\mathcal{L}_{\mathrm{JSD}} \\
  &\quad + \lambda_{\mathrm{Margin}}\mathcal{L}_{\mathrm{Margin}}
   + \lambda_{\mathrm{Sum}}\mathcal{L}_{\mathrm{Sum}},
\end{aligned}
\end{equation*}
where $\lambda_{\mathrm{LM}}$, $\lambda_{\mathrm{JSD}}$, $\lambda_{\mathrm{Margin}}$, and $\lambda_{\mathrm{Sum}}$ are hyperparameters, all set to 1 for simplicity.

\section{Experimental Settings}
\label{sec:settings}

\paragraph{Models}
We employ three open-weight LLMs: \textsc{Llama-3.1-8B-Instruct} \cite{grattafiori2024llama3herdmodels}, \textsc{Mistral-7B-Instruct-v0.3} \cite{jiang2023mistral7b}, and \textsc{Gemma-2-9b-it} \cite{gemmateam2024gemma2improvingopen}.

\paragraph{Datasets}
We conduct experiments across three open-ended QA datasets: TriviaQA \cite{joshi-etal-2017-triviaqa}, MMLU \cite{hendrycks2021measuring}, and LogiQA \cite{10.5555/3491440.3491941}. 
Notably, we train only on TriviaQA, enabling evaluation of out-of-distribution (OOD) generalization.

\paragraph{Confidence verbalization types}
Following \citet{yang2025on}, we adopt five verbalization types: 
\begin{itemize}[itemsep=0pt, topsep=5pt, leftmargin=13pt]
    \item \textbf{ScoreLetter}: letter grades (\{\texttt{E}, \texttt{D}, \texttt{C}, \texttt{B}, \texttt{A}\}).
    \item \textbf{ScoreNumber}: integer scores (\{\texttt{0}, \texttt{1}, \dots, \texttt{9}\}).
    \item \textbf{ScoreText}: categories (\{\texttt{low}, \texttt{medium}, \texttt{high}\}).
    \item \textbf{ScoreFloat}: floating-point values (\{\texttt{0.0}--\texttt{1.0}\}).
    \item \textbf{ScorePercent}: percentages (\{\texttt{0}, \texttt{1}, \dots, \texttt{100}\}).
\end{itemize}
Confidence expressions are ordered in ascending magnitude, with later ones denoting higher confidence.
During training, ADVICE uses \textbf{ScoreLetter} and \textbf{ScoreNumber}; at inference, all scoring types are employed as required by the specific evaluation.
More details are provided in Appendix \ref{appendix:confidence_types}.

\paragraph{Baselines}
We also compare against several confidence estimation methods: (1) \textbf{Default}, which refers to the naïve use of LLMs with minimal prompting; (2) \textbf{Prompting}, which augments the Default approach with explicit instructions to consider the self-generated answer; (3) \textbf{Self-Consistency} \cite{xiong2024can}, a sampling-based approach that generates multiple verbalized confidence scores and aggregates them; and (4) \textbf{ConfTuner} \cite{li2025conftunertraininglargelanguage}, which fine-tunes LLMs to align confidence distributions with empirical correctness by optimizing tokenized Brier scores.

\paragraph{Metrics}

We evaluate confidence calibration quality with 4 metrics:
Expected Calibration Error (\textbf{ECE}) \cite{Pakdaman_Naeini_Cooper_Hauskrecht_2015}, absolute Net Calibration Error (\textbf{|NCE|}) \cite{groot-valdenegro-toro-2024-overconfidence}, Brier score (\textbf{BS}) \cite{VERIFICATIONOFFORECASTSEXPRESSEDINTERMSOFPROBABILITY}, and Area Under the ROC Curve (\textbf{AUROC}) \cite{10.1007/978-3-642-40994-3_29}.
Lower values are better for the first three metrics, while higher values are better for AUROC.
Appendix \ref{appendix:experimental_setting} provides further details.

\section{Experimental Results}

\begin{table*}[t]   
\centering
\scriptsize
\setlength{\tabcolsep}{1.3pt}
\renewcommand{\arraystretch}{1.1}
\begin{tabular}{llllllllllllll}
\toprule
\multirow{2}[2]{*}{\textbf{Model}} & \multirow{2}[2]{*}{\textbf{Method}} & \multicolumn{4}{c}{\textbf{TriviaQA}} & \multicolumn{4}{c}{\textbf{MMLU}} & \multicolumn{4}{c}{\textbf{LogiQA}} \\ 

\cmidrule(lr){3-6}\cmidrule(lr){7-10}\cmidrule(lr){11-14}

& & {ECE ($\downarrow$)} & |NCE| ($\downarrow$) & BS ($\downarrow$) & AUROC ($\uparrow$) & ECE & |NCE| & BS & AUROC& ECE & |NCE| & BS & AUROC\\ 

\midrule

\multirow{6}{*}{\shortstack[l]{\textsc{Llama3.1} \\ \textsc{8B} \\ \textsc{Instruct}}}  & Default & 16.9 & 16.6 & 21.2 & 56.2 & 26.9 & 26.7 & 29.7 & 50.8 & 53.8 & 53.3 & 52.9 &50.5 \\ 
& Prompting & 12.1 & 12.0 & 17.6 & 59.7 & 25.5 & 25.3 & 28.5 & 52.8 & 48.1 & 47.4 & 48.0 & 50.4 \\
& Self-Consistency & 15.7$_{\!\pm 0.1}$ & 15.1$_{\pm 0.2}$ & 22.2$_{\pm 0.1}$ & 58.6$_{\pm 0.9}$ & 25.0$_{\pm 1.5}$ & 25.0$_{\pm 1.5}$ & 29.9$_{\pm 1.1}$ & 53.9$_{\pm 0.8}$ & 45.3$_{\pm 1.0}$ & 45.1$_{\pm 1.1}$ & 46.2$_{\pm 0.7}$ & 45.6$_{\pm 1.3}$ \\ 
& ConfTuner & \bfseries \phantom{0}5.2 & \bfseries \phantom{0}1.1 & 15.3 & 66.3 & 13.9 & 13.9 & 24.2 & 58.2 & 28.6 & 28.2 & 32.5 & 54.4 \\ 
\addlinespace[0.8pt]
& \cellcolor{gray!25}\textbf{ADVICE (Ours)} & \cellcolor{gray!25}10.4$_{\pm 1.2}$ & \cellcolor{gray!25}\phantom{0}9.8$_{\pm 1.1}$ & \cellcolor{gray!25}\bfseries 14.8$_{\pm 0.5}$ & \cellcolor{gray!25}\underline{77.0}$_{\pm 0.1}$ & \cellcolor{gray!25}\bfseries \phantom{0}8.6$_{\pm 2.3}$ & \cellcolor{gray!25}\phantom{0}\underline{7.6}$_{\pm 0.7}$ & \cellcolor{gray!25}\bfseries 20.7$_{\pm 0.7}$ & \cellcolor{gray!25}\bfseries 69.2$_{\pm 0.9}$ & \cellcolor{gray!25}\bfseries 23.0$_{\pm 2.7}$ & \cellcolor{gray!25}\bfseries 21.2$_{\pm 2.9}$ & \cellcolor{gray!25}\bfseries 30.1$_{\pm 1.9}$ & \cellcolor{gray!25}\underline{57.9}$_{\pm 0.7}$ \\
& \cellcolor{gray!25}\quad \textit{w/} ConfTuner & \cellcolor{gray!25}\phantom{0}\underline{9.4}$_{\pm0.4}$ & \cellcolor{gray!25}\phantom{0}\underline{8.7}$_{\pm0.4}$ & \cellcolor{gray!25}\underline{15.1}$_{\pm0.1}$ & \cellcolor{gray!25}\bfseries77.9$_{\pm0.1}$ & \cellcolor{gray!25}\phantom{0}\underline{9.6}$_{\pm0.9}$ & \cellcolor{gray!25}\phantom{0}\bfseries7.1$_{\pm1.4}$ & \cellcolor{gray!25}\underline{20.9}$_{\pm0.4}$ & \cellcolor{gray!25}\underline{68.7}$_{\pm0.8}$ & \cellcolor{gray!25}\underline{26.0}$_{\pm0.7}$ & \cellcolor{gray!25}\underline{23.8}$_{\pm0.9}$ & \cellcolor{gray!25}\underline{31.1}$_{\pm0.4}$ & \cellcolor{gray!25}\bfseries 58.6$_{\pm0.3}$ \\ 

\midrule

\multirow{6}{*}{\shortstack[l]{\textsc{Mistral} \\ \textsc{7B} \\ \textsc{Instruct}}} & Default & 32.8 & 30.5 & 35.3 & 51.6 & 35.3 & 35.0 & 37.0 & 51.5 & 51.8 & 50.9 & 51.3 &52.1 \\ 
& Prompting & 27.6 & 24.8 & 31.0 & 52.8 & 36.3 & 34.1 & 38.2 & 49.9 & 49.2 & 48.0 & 49.2 & 51.4 \\
& Self-Consistency & 31.3$_{\pm 0.4}$ & 30.3$_{\pm 0.5}$ & 31.6$_{\pm 0.3}$ & 68.5$_{\pm 1.0}$ & 34.5$_{\pm 0.5}$ & 34.3$_{\pm 0.3}$ & 36.1$_{\pm 0.3}$ & 61.0$_{\pm 0.8}$ & 45.7$_{\pm 0.8}$ & 45.4$_{\pm 0.6}$ & 43.2$_{\pm 1.0}$ & 59.0$_{\pm 2.0}$ \\ 

& ConfTuner & \phantom{0}\underline{8.1}$_{\pm 1.9}$ & \phantom{0}\underline{7.5}$_{\pm 2.3}$ & \textbf{16.3}$_{\pm 0.5}$ & \textbf{80.3}$_{\pm 0.4}$ & 36.0$_{\pm 2.1}$ & 36.0$_{\pm 2.1}$ & 34.3$_{\pm 1.3}$ & \textbf{70.8}$_{\pm 1.6}$ & \underline{24.8}$_{\pm 2.3}$ & \underline{24.8}$_{\pm 2.3}$ & \textbf{22.1}$_{\pm 2.3}$ & \underline{63.1}$_{\pm 1.1}$ \\ 
\addlinespace[0.8pt]
& \cellcolor{gray!25}\textbf{ADVICE (Ours)} & \cellcolor{gray!25}14.5$_{\pm 2.3}$ & \cellcolor{gray!25}\phantom{0}8.0$_{\pm 2.4}$ & \cellcolor{gray!25}20.6$_{\pm 1.2}$ & \cellcolor{gray!25}74.7 $_{\pm 2.1}$ & \cellcolor{gray!25}\underline{28.7}$_{\pm 2.5}$ & \cellcolor{gray!25}\underline{27.3}$_{\pm 2.9}$ & \cellcolor{gray!25}\underline{31.4}$_{\pm 1.9}$ & \cellcolor{gray!25}62.7$_{\pm 3.1}$ & \cellcolor{gray!25}35.6$_{\pm 5.5}$ & \cellcolor{gray!25}31.6$_{\pm 6.1}$ & \cellcolor{gray!25}38.2$_{\pm 4.3}$ & \cellcolor{gray!25}59.3$_{\pm 1.7}$ \\ 
&  \cellcolor{gray!25}\quad \textit{w/} ConfTuner & \cellcolor{gray!25}\bfseries \phantom{0}6.9$_{\pm 1.2}$ & \cellcolor{gray!25}\phantom{0}\textbf{4.7}$_{\pm 1.4}$ & \cellcolor{gray!25}\underline{17.8}$_{\pm 1.1}$ & \cellcolor{gray!25}\underline{76.4}$_{\pm 1.4}$ & \cellcolor{gray!25}\textbf{21.8}$_{\pm 0.4}$ & \cellcolor{gray!25}\textbf{21.3}$_{\pm 0.5}$ & \cellcolor{gray!25}\textbf{26.5}$_{\pm 0.1}$ & \cellcolor{gray!25}\underline{69.9}$_{\pm 0.3}$ & \cellcolor{gray!25}\textbf{24.0}$_{\pm 1.0}$ & \cellcolor{gray!25}\textbf{23.1}$_{\pm 1.1}$ & \cellcolor{gray!25}\underline{28.1}$_{\pm 0.4}$ & \cellcolor{gray!25}\textbf{68.0}$_{\pm 0.4}$ \\

\midrule

\multirow{6}{*}{\shortstack[l]{\textsc{Gemma2} \\ \textsc{9b-it}}}  & Default & 21.9 & 21.8 & 25.3 & 52.7 & 21.0 & 21.0 & 24.7 & 50.1 & 39.1 & 39.0 & 40.4 & 50.9 \\ 

& Prompting & 21.4 & 21.3 & 24.9 & 53.9 & 21.0 & 21.0 & 24.6 & 50.4 & 39.0 & 38.7 & 40.3 & 50.3 \\

& Self-Consistency & 28.5$_{\pm 0.5}$ & 28.2$_{\pm 0.4}$ & 28.5$_{\pm 0.5}$ & 65.2$_{\pm 1.1}$ & 22.5$_{\pm 0.4}$ & 21.8$_{\pm 0.4}$ & 25.8$_{\pm 0.2}$ & 57.4$_{\pm 0.9}$ & 39.0$_{\pm 0.5}$ & 38.8$_{\pm 0.5}$ & 41.8$_{\pm 0.4}$ & 44.8$_{\pm 0.5}$ \\ 

& ConfTuner & \phantom{0}\underline{5.7}$_{\pm 0.4}$ & \phantom{0}\underline{2.9}$_{\pm 1.5}$ & \textbf{14.3}$_{\pm 0.3}$ & \textbf{82.7}$_{\pm 0.2}$ & \underline{11.0}$_{\pm 1.0}$ & \phantom{0}\underline{7.3}$_{\pm 0.9}$ & 20.2$_{\pm 0.3}$ & \textbf{75.7}$_{\pm 0.4}$ & 18.4$_{\pm 1.1}$ & 17.9$_{\pm 1.1}$ & \textbf{23.8}$_{\pm 0.4}$ & \textbf{71.4}$_{\pm 0.3}$ \\ 

\addlinespace[0.8pt]

& \cellcolor{gray!25}\textbf{ADVICE (Ours)} & \cellcolor{gray!25}\phantom{0}6.2$_{\pm 3.2}$ & \cellcolor{gray!25}\phantom{0}5.1$_{\pm 3.8}$ & \cellcolor{gray!25}16.3$_{\pm 0.5}$ & \cellcolor{gray!25}\underline{77.4}$_{\pm 0.5}$ & \cellcolor{gray!25}\phantom{0}\textbf{5.6}$_{\pm 0.4}$ & \cellcolor{gray!25}\phantom{0}\textbf{3.9}$_{\pm 1.0}$ & \cellcolor{gray!25}\textbf{18.6}$_{\pm 0.3}$ & \cellcolor{gray!25}65.9$_{\pm 1.4}$ & \cellcolor{gray!25}\underline{11.9}$_{\pm 2.6}$ & \cellcolor{gray!25}\underline{10.9}$_{\pm 2.8}$ & \cellcolor{gray!25}25.8$_{\pm 0.3}$ & \cellcolor{gray!25}57.8$_{\pm 0.2}$\\

& \cellcolor{gray!25}\quad \textit{w/} ConfTuner & \cellcolor{gray!25}\phantom{0}\textbf{3.4}$_{\pm 0.6}$ & \cellcolor{gray!25}\phantom{0}\textbf{2.0}$_{\pm 1.2}$ & \cellcolor{gray!25}\underline{16.0}$_{\pm 0.2}$ & \cellcolor{gray!25}77.1$_{\pm 1.0}$ & \cellcolor{gray!25}11.7$_{\pm 2.6}$ & \cellcolor{gray!25}11.7$_{\pm 2.6}$ & \cellcolor{gray!25}\underline{19.9}$_{\pm 1.0}$ & \cellcolor{gray!25}\underline{66.8}$_{\pm 1.9}$ & \cellcolor{gray!25}\phantom{0}\textbf{8.0}$_{\pm 0.3}$ & \cellcolor{gray!25}\phantom{0}\textbf{6.2}$_{\pm 0.2}$ & \cellcolor{gray!25}\underline{24.6}$_{\pm 0.1}$ & \cellcolor{gray!25}\underline{69.0}$_{\pm 0.3}$\\ 
\bottomrule
\end{tabular}
\caption{Average performance across two seen verbalization types (i.e., Score\{Letter, Number\}) and three random seeds, with standard deviations. Evaluation is conducted on in-domain (TriviaQA) and out-of-distribution (MMLU and LogiQA) datasets. Best results are in \textbf{bold}, and second-best results are \underline{underlined}. All values are reported as percentages. ADVICE matches ConfTuner and yields orthogonal gains when combined.}
\label{tab:main_table}
\end{table*}

\begin{figure*}[t]
\centering
\begin{subfigure}{0.24\textwidth}
  \centering
  \includegraphics[width=\linewidth]{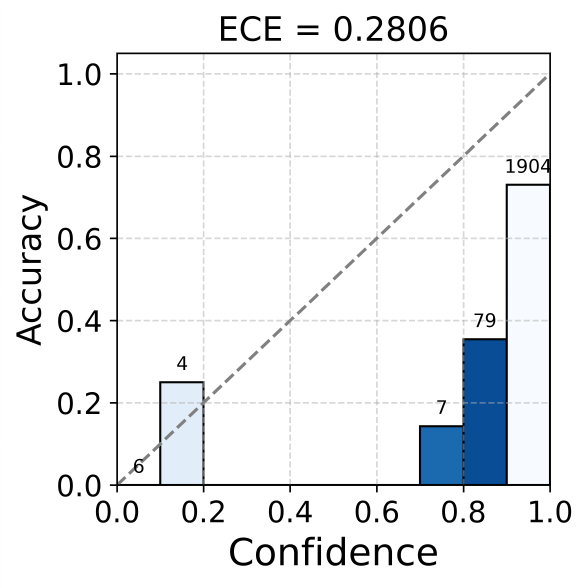}
  \caption{Prompting}
\end{subfigure} \hfill
\begin{subfigure}{0.24\textwidth}
  \centering
  \includegraphics[width=\linewidth]{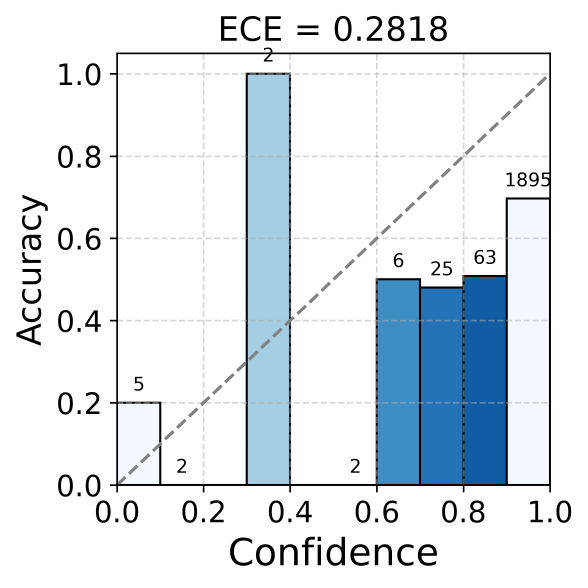}
  \caption{Self-Consistency}
\end{subfigure} \hfill
\begin{subfigure}{0.24\textwidth}
  \centering
  \includegraphics[width=\linewidth]{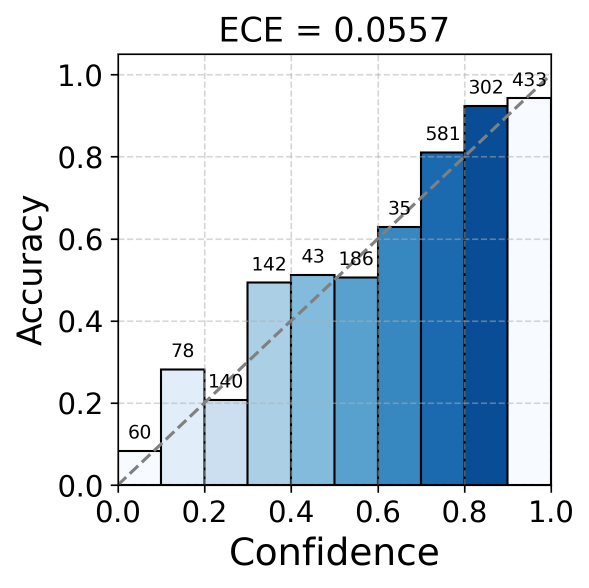}
  \caption{ConfTuner}
\end{subfigure} \hfill
\begin{subfigure}{0.24\textwidth}
  \centering
  \includegraphics[width=\linewidth]{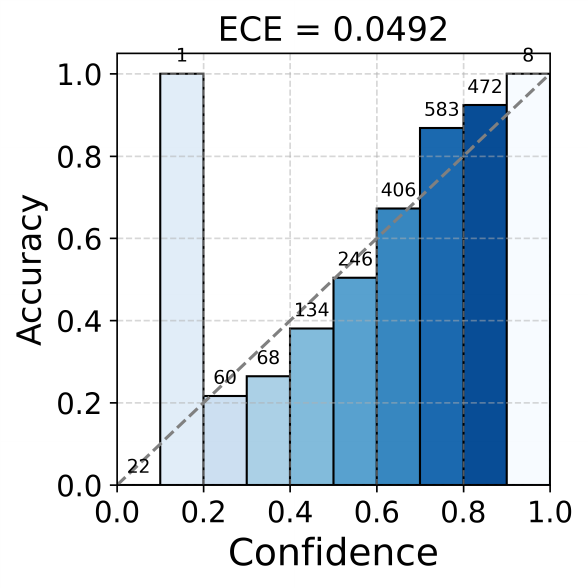}
  \caption{ADVICE}
\end{subfigure}
\caption{Reliability diagrams on TriviaQA with \textsc{Gemma-2-9b-it} under the ScoreNumber setting, where numbers above each bin indicate the number of data instances. ADVICE achieves high calibration quality comparable to ConfTuner, demonstrating their effectiveness. Additional cases are illustrated in Figure \ref{fig:reliability_diagram_appendix1}, \ref{fig:reliability_diagram_appendix2} in the Appendix.}
\label{fig:reliability_diagram}
\end{figure*}

\subsection{Main Results}
\label{sec:main_results}

Table~\ref{tab:main_table} summarizes our main experimental results, from which we derive three key findings.

\paragraph{ADVICE leads to improved confidence calibration with strong OOD generalization.}
When evaluated on TriviaQA, which is used for both training and evaluation, ADVICE consistently outperforms the Default, Prompting, and Self-Consistency baselines, effectively mitigating LLM overconfidence. 
Compared to ConfTuner, ADVICE achieves comparable performance across diverse settings, supporting the viability of training-based approaches.
Beyond aggregate metrics, Figure \ref{fig:reliability_diagram} provides qualitative evidence of this advantage: while Prompting and Self-Consistency produce uniformly high confidence scores with limited reliability, ADVICE yields fine-grained confidence estimates that more closely track accuracy, resulting in more precise predictions.

Furthermore, as training inherently carries a risk of overfitting, robustness to domain shift is essential for training-based methods. 
We therefore evaluate ADVICE and ConfTuner under OOD settings using MMLU and LogiQA. 
ADVICE outperforms ConfTuner in 19 of 24 cases (2 datasets × 4 metrics × 3 models), demonstrating strong robustness as a general-purpose confidence calibration framework.

\paragraph{ADVICE provides orthogonal gains over ConfTuner.}
Table~\ref{tab:main_table} further shows that combining ADVICE with ConfTuner (i.e., \textit{w/} ConfTuner) often yields extra gains, implying that the two methods tackle orthogonal aspects of the limitations in verbalized confidence estimation.
ConfTuner directly aligns confidence token generation with answer predictions using existing datasets, which may make it susceptible to overfitting.
In contrast, ADVICE, while also based on fine-tuning, guides the model to condition confidence estimation more strongly on its own generated answers, indirectly improving confidence calibration. 
As a result, ADVICE offers greater robustness under distribution shift, complementing ConfTuner’s shortcomings.
Appendix \ref{appendix:implementation_details} provides details of combining two methods.

\begin{figure}[t]
\centering
\includegraphics[width=0.99\columnwidth]{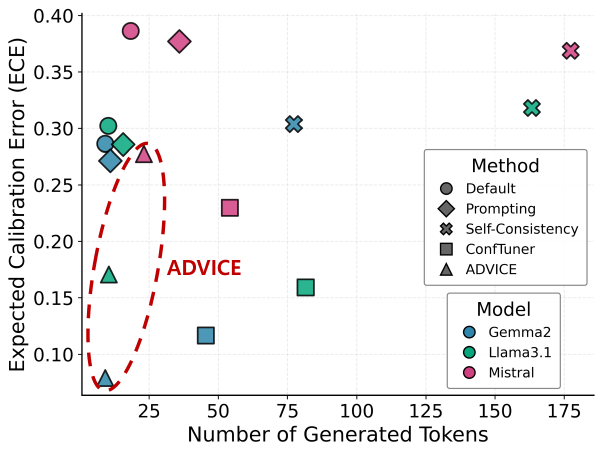}
\caption{Calibration performance (ECE)–efficiency (token usage) across method–model pairs. 
Compared to the baselines, ADVICE achieves the best balance, requiring the lowest generation cost while maintaining reliable confidence estimation.}
\label{fig:efficiency}
\end{figure}

\paragraph{ADVICE achieves the best trade-off between performance and efficiency.}
Figure \ref{fig:efficiency} visualizes the performance–efficiency relationship for confidence estimation, where performance is measured by ECE and efficiency by the number of generated tokens (i.e., token usage).\footnote{Token usage is computed as the total number of tokens generated for the answer and the verbalized confidence.}
For both metrics, lower values are better, making the lower-left region ideal. 
Each point denotes performance averaged over diverse prompt types, datasets, and random seeds. 

Across all three LLMs, Default and Prompting are relatively cost-efficient but consistently exhibit high ECE, indicating limited calibration quality. 
Self-Consistency dramatically increases token usage---reflecting its multi-sampling nature---yet still delivers unsatisfactory performance.
On the other hand, ConfTuner improves calibration over these decoding-based baselines but typically incurs higher token usage for confidence verbalization.
Notably, ADVICE clusters in the lower-left region, achieving lower ECE with fewer generated tokens and thus offering a more favorable balance between confidence calibration performance and efficiency.

\begin{table}[t]
\centering
\scriptsize
\setlength{\tabcolsep}{2.5pt}
\renewcommand{\arraystretch}{1.0}
\begin{tabular}{llrrrrrr}
\toprule
\multirow{2}[2]{*}{\textbf{Model}} & \multirow{2}[2]{*}{\textbf{Training Obj.}} & \multicolumn{3}{c}{\textbf{TriviaQA}} & \multicolumn{3}{c}{\textbf{MMLU}} \\ 
\addlinespace[-1pt]
\cmidrule(lr){3-5}\cmidrule(lr){6-8}
\addlinespace[-1pt]
& & ECE & |NCE| & BS & ECE & |NCE| & BS  \\ 
\midrule
\multirow{8}{*}{\shortstack[l]{\textsc{Gemma2} \\ \textsc{9b-it}}}
& LM & 23.0 & 23.0 & 25.6 & 22.5 & 22.5 & 25.2 \\
& LM+JSD & \underline{8.6} & \underline{1.1} & \underline{16.7} & 13.2 & 11.7 & 20.5 \\ 
& LM+Margin & 16.8 & \textbf{0.9} & 20.2 & 21.9 & 13.1 & 24.7 \\ 
& LM+Sum & 21.1 & 21.1 & 24.2 & 19.5 & 19.5 & 23.9 \\ 
& LM+JSD+Margin & 11.0 & 4.0 & 17.3 & 14.3 & 11.3 & 21.0 \\ 
& LM+JSD+Sum & 15.3 & 14.8 & 18.9 & \underline{7.5} & \underline{7.1} & \underline{19.0} \\ 
& LM+Margin+Sum & 20.9 & 20.2 & 23.7 & 19.7 & 19.7 & 23.8 \\
\addlinespace[-1.5pt]
\cmidrule{2-8}
\addlinespace[-1.5pt]
& \textbf{ADVICE} & \textbf{6.2} & 5.1 & \textbf{16.3} & \textbf{5.6} & \textbf{0.4} & \textbf{18.6} \\ 
\midrule
\multirow{8}{*}{\shortstack[l]{\textsc{Llama3.1} \\ \textsc{8B Instruct}}}
& LM & 13.3 & 13.3 & 17.3 & 27.7 & 27.7 & 29.9 \\
& LM+JSD & \textbf{6.2} & \textbf{3.4} & \textbf{14.2} & 20.5 & 20.1 & 25.1 \\ 
& LM+Margin & 17.4 & \underline{6.8} & 20.6 & 28.2 & 24.6 & 30.2 \\ 
& LM+Sum & 32.5 & 32.5 & 26.8 & 20.3 & 20.3 & 26.0 \\ 
& LM+JSD+Margin & 11.0 & 7.4 & 16.8 & 23.3 & 19.3 & 26.4 \\ 
& LM+JSD+Sum & 19.5 & 19.5 & 17.8 & \textbf{5.5} & \textbf{2.0} & \textbf{19.8} \\ 
& LM+Margin+Sum & 31.0 & 31.0 & 25.3 & 15.1 & 15.1 & 23.0 \\
\addlinespace[-1.5pt]
\cmidrule{2-8}
\addlinespace[-1.5pt]
& \textbf{ADVICE} & \underline{10.4} & 9.8 & \underline{14.8} & \underline{8.6} & \underline{7.6} & \underline{20.7} \\
\bottomrule
\end{tabular}
\caption{Ablation study of the final training objective. All values are reported as percentages. Best and second-best results are indicated in \textbf{bold} and \underline{underlined}.}
\label{tab:loss_ablation}
\end{table}

\subsection{Ablation Study on Training Objectives}
We conduct an ablation study on the final training objective to assess the contribution of its components.
First, as described in \S\ref{sec:training_objective}, $\mathcal{L}_{\mathrm{LM}}$ serves as an auxiliary term for preserving language modeling performance.
As a result, we observe that $\mathcal{L}_{\mathrm{LM}}$ is generally independent of confidence calibration.
Second, the results in Table \ref{tab:loss_ablation} show that $\mathcal{L}_{\mathrm{JSD}}$ and $\mathcal{L}_{\mathrm{Margin}}$ contribute to improving confidence verbalization together. 
Compared to using either loss alone, jointly optimizing both enables the model not only to distinguish $P_{\text{correct}}$ from $P_{\text{wrong}}$ but also to separate them in the intended direction.
$\mathcal{L}_\mathrm{Sum}$ enforces the definition of confidence, resulting in improved calibration on OOD benchmarks.

Beyond individual components, we further examine the contribution of their combinations.
While $\mathcal{L}_{\mathrm{JSD}}$ alone tends to favor in-domain performance, $\mathcal{L}_{\mathrm{Margin}}$ and $\mathcal{L}_{\mathrm{Sum}}$ together exhibit the opposite tendency, promoting coarse separation between correct and incorrect answers yet failing to adequately regulate the overall confidence distribution.
Overall, ADVICE combines these terms to achieve robust generalization across datasets and models.

\begin{figure*}[t!]
\centering
\begin{subfigure}{0.24\textwidth}
\centering
\includegraphics[width=\linewidth]{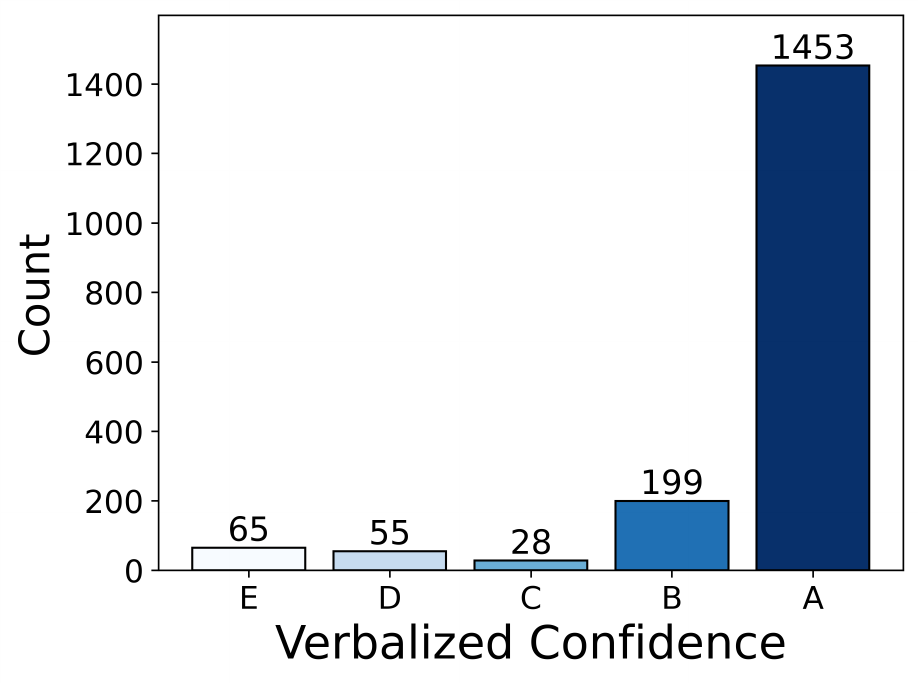}
\caption{\centering \textsc{Llama3.1} \\ (TriviaQA, Default)}
\label{fig:answer_masking_default1}
\end{subfigure} \hfill
\begin{subfigure}{0.24\textwidth}
\centering
\includegraphics[width=\linewidth]{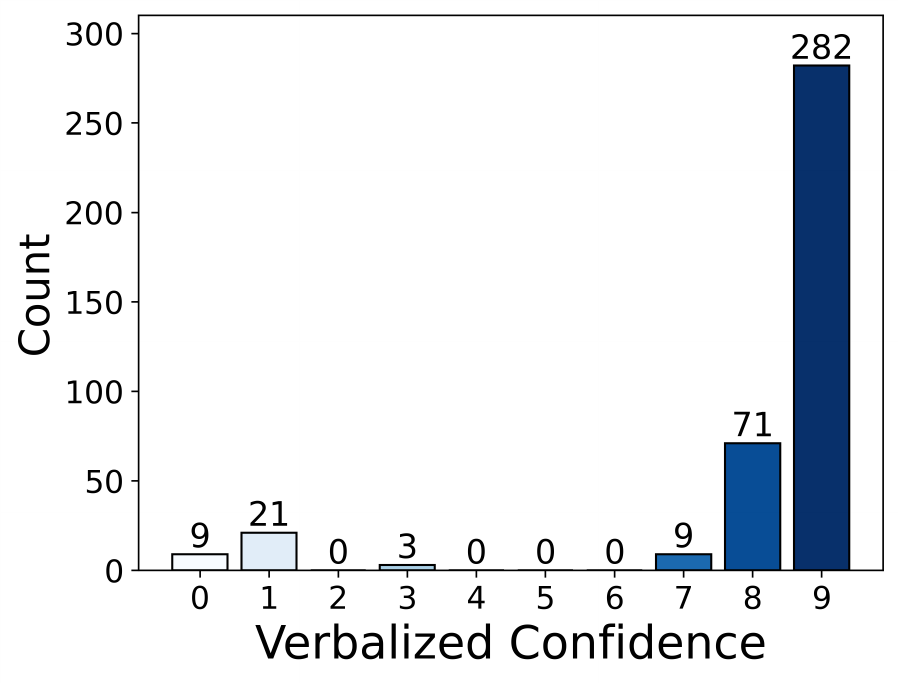}
\caption{\centering \textsc{Mistral} \\ (LogiQA, Default)}
\label{fig:answer_masking_default2}
\end{subfigure}\hfill
\begin{subfigure}{0.24\textwidth}
\centering
\includegraphics[width=\linewidth]{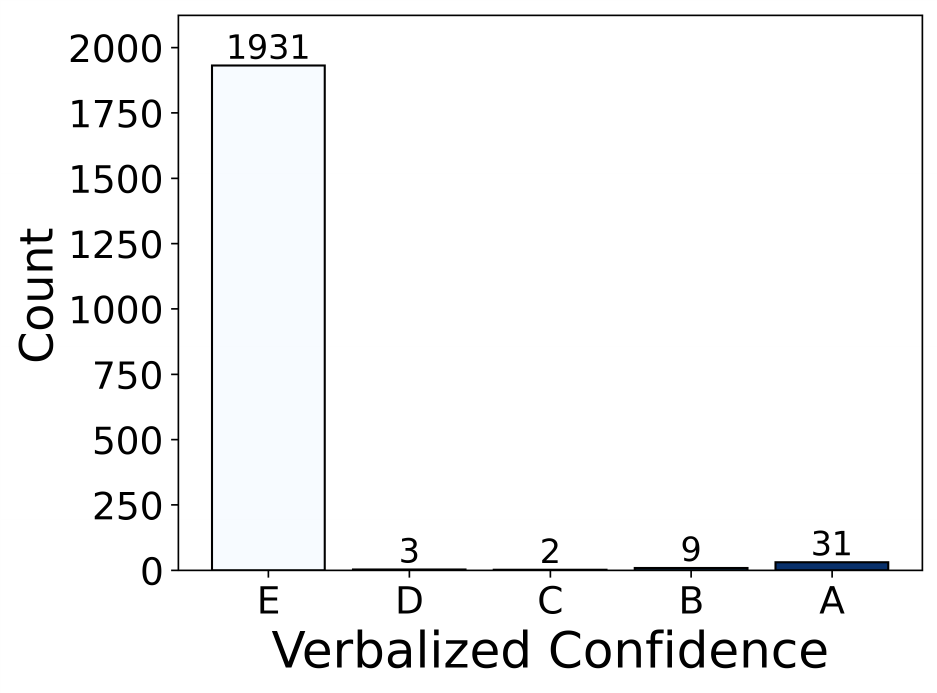}
\caption{\centering \textsc{Llama3.1} \\ (TriviaQA, ADVICE)}
\label{fig:answer_masking_advice1}
\end{subfigure} \hfill
\begin{subfigure}{0.24\textwidth}
\centering
\includegraphics[width=\linewidth]{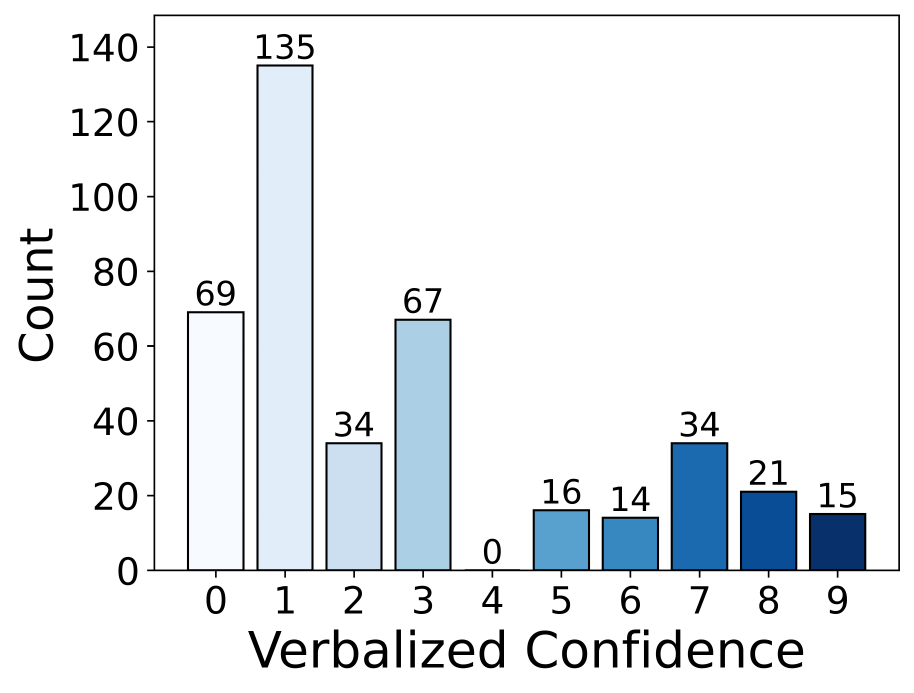}
\caption{\centering \textsc{Mistral} \\ (LogiQA, ADVICE)}
\label{fig:answer_masking_advice2}
\end{subfigure}
\caption{Verbalized confidence distributions after answer masking. Default remains overconfident without answers, whereas ADVICE reallocates probability mass toward less confident expressions (e.g., E, 0, and 1).}
\label{fig:answer_masking}
\end{figure*}

\subsection{Generalization on Verbalization Types}
As explained in \S\ref{sec:settings}, we construct the training set for ADVICE using two verbalization types (ScoreLetter and ScoreNumber).
We then evaluate ADVICE on three other formats---ScoreText, ScoreFloat, and ScorePercent---to validate its robustness across verbalization schemes.
Table \ref{tab:prompt_generalization} reports performance on these unseen types, demonstrating consistent results across formats.
Together with strong OOD performance in \S \ref{sec:main_results}, these results indicate that ADVICE learns a transferable expression of confidence rather than exploiting in-domain shortcuts.

\begin{table}[t]
\centering
\scriptsize
\setlength{\tabcolsep}{2.5pt}
\begin{tabular}{lllrrrrrr}
\toprule
\multirow{2}[2]{*}{\textbf{Model}} & \multirow{2}[2]{*}{\shortstack[l]{\textbf{Verb.} \\ \textbf{Type}}} & \multirow{2}[2]{*}{\textbf{Method}} & \multicolumn{3}{c}{\textbf{TriviaQA}} & \multicolumn{3}{c}{\textbf{MMLU}} \\ 
\cmidrule(lr){4-6}\cmidrule(lr){7-9}
& & & ECE & |NCE| & BS & ECE & |NCE| & BS  \\ 
\midrule
\multirow{7}{*}{\shortstack[l]{\textsc{Gemma2} \\ \textsc{9b-it}}} 
& \multirow{2}{*}{Text}
& Default & 17.1 & 17.1 & 22.3 & 25.5 & 25.5 & 26.5 \\ 
& & \textbf{ADVICE} & \textbf{8.3} & \textbf{6.7} & \textbf{17.6} & \textbf{6.8} & \textbf{0.9} & \textbf{18.4} \\ 
\cmidrule{2-9}
& \multirow{2}{*}{Percent}
& Default & 26.9 & 26.8 & 27.0 & 25.5 & 25.5 & 26.5 \\ 
& & \textbf{ADVICE} & \textbf{6.7} & \textbf{0.9} & \textbf{16.4} & \textbf{8.9} & \textbf{8.7} & \textbf{19.2} \\ 
\cmidrule{2-9}
& \multirow{2}{*}{Float}
& Default & 27.5 & 27.4 & 27.3 & 26.3 & 26.3 & 27.0 \\
& & \textbf{ADVICE} & \textbf{6.2} & \textbf{4.9} & \textbf{15.4} & \textbf{9.2} & \textbf{8.7} & \textbf{19.2} \\ 
\midrule
\multirow{6}{*}{\shortstack[l]{\textsc{Llama3.1} \\ \textsc{8B Instruct}}} 
& \multirow{2}{*}{Text}
& Default & 26.9 & 26.8 & 27.0 & 25.5 & 25.5 & 26.5 \\ 
& & \textbf{ADVICE} & \textbf{11.9} & \textbf{11.9} & \textbf{16.4} & \textbf{8.0} & \textbf{3.7} & \textbf{21.5} \\ 
\cmidrule{2-9}
& \multirow{2}{*}{Percent}
& Default & 18.6 & 18.5 & 20.3 & 30.2 & 30.2 & 31.6 \\ 
& & \textbf{ADVICE} & \textbf{5.3} & \textbf{4.1} & \textbf{14.0} & \textbf{13.2} & \textbf{12.9} & \textbf{22.7} \\ 
\cmidrule{2-9}
& \multirow{2}{*}{Float}
& Default & 19.6 & 19.5 & 21.0 & 32.5 & 32.3 & 33.0 \\
& & \textbf{ADVICE} & \textbf{5.7} & \textbf{5.7} & \textbf{14.1} & \textbf{14.9} & \textbf{14.7} & \textbf{22.3} \\ 
\bottomrule
\end{tabular}
\caption{Performance on unseen verbalization formats: ScoreText, ScorePercent, and ScoreFloat. All values are percentages. The best results are in \textbf{bold}.}
\label{tab:prompt_generalization}
\end{table}

\subsection{Effect on General Task Performance}
\label{sec:general_performance}
When fine-tuning an LLM, it is essential to verify that the modification does not compromise general task performance. 
Accordingly, we examine the impact of ADVICE on task (QA) accuracy.
As shown in Table \ref{tab:accuracy}, accuracy changes are negligible, meaning that ADVICE preserves the LLM’s capabilities. 
Since verbalized confidence is often overconfident, increases in task accuracy could affect calibration metrics such as ECE even if confidence estimation remains unchanged. 
However, stable accuracy after fine-tuning suggests that the observed ECE reductions stem from improved confidence calibration rather than gains in task performance.

\section{ADVICE Enhances Answer Awareness}
Ultimately, we conduct analyses to provide evidence that ADVICE’s improvements originate from its answer-groundedness. 
To this end, we introduce a new experiment and revisit the analyses presented in \S\ref{sec:Claim: Verbalized Confidence is Nearly Answer-Independent} after training LLMs with ADVICE.

\begin{table}[t]
\centering
\scriptsize
\setlength{\tabcolsep}{1.5pt}
\begin{tabular}{llllll}
\toprule
\multirow{2}[2]{*}{\textbf{Dataset}} & \multirow{2}[2]{*}{\textbf{Method}} & \multicolumn{2}{c}{\textsc{Gemma2-9b-it}} & \multicolumn{2}{c}{\textsc{Llama3.1-8B-Instruct}} \\ 
\cmidrule(lr){3-4}\cmidrule(lr){5-6}
& & ScoreLetter & ScoreNumber & ScoreLetter & ScoreNumber \\ 
\midrule
\multirow{2}{*}{\textbf{TriviaQA}}
& Default & \phantom{0}70.7 & \phantom{0}70.7 & \phantom{0}75.2 & \phantom{0}74.8 \\ 
& \textbf{ADVICE} & \phantom{0}71.6$_{\pm 0.2}$ & \phantom{0}71.7$_{\pm 0.4}$& \phantom{0}78.1$_{\pm 0.1}$ & \phantom{0}77.4$_{\pm 0.5}$ \\ 
\midrule
\multirow{2}{*}{\textbf{MMLU}}
& Default & \phantom{0}72.2 & \phantom{0}72.1 & \phantom{0}66.6 & \phantom{0}67.2 \\ 
& \textbf{ADVICE} & \phantom{0}73.0$_{\pm 0.1}$& \phantom{0}72.8$_{\pm 0.3}$& \phantom{0}67.1$_{\pm 0.3}$ & \phantom{0}67.5$_{\pm 0.76}$ \\ 
\midrule
\multirow{2}{*}{\textbf{LogiQA}}
& Default & \phantom{0}53.2 & \phantom{0}52.3 & \phantom{0}39.0 & \phantom{0}38.9 \\ 
& \textbf{ADVICE} & \phantom{0}52.8$_{\pm 0.2}$ & \phantom{0}52.5$_{\pm 0.1}$ & \phantom{0}41.9$_{\pm 0.2}$ & \phantom{0}41.6$_{\pm 0.1}$ \\ 
\bottomrule
\end{tabular}
\caption{Task (QA) accuracies before and after fine-tuning. These results demonstrate that ADVICE does not adversely impact the task performance of LLMs.}
\label{tab:accuracy}
\end{table}

In the first study, we replace answer tokens with an equal-length sequence of padding (e.g., \texttt{<pad>}) tokens to simulate the absence of the answer and evaluate its effect.
As illustrated in Figures \ref{fig:answer_masking_default1} and \ref{fig:answer_masking_default2}, the Default method (i.e., before training) produces confidence distributions that are markedly skewed toward high values, indicating overconfidence.
In contrast, ADVICE (Figures \ref{fig:answer_masking_advice1} and \ref{fig:answer_masking_advice2}; after training) reveals the opposite behavior: its verbalized confidence substantially declines when the answer is masked, conveying obscurity regarding the correctness of the response.
This finding empirically validates that ADVICE enhances the model’s answer-awareness in confidence estimation.

\begin{table*}[t!]
\centering
\scriptsize
\setlength{\tabcolsep}{5pt}
\begin{tabular}{l*{10}{c}}
\toprule
\textbf{Training Step} & \multicolumn{10}{c}{\textbf{Top 10 Tokens}} \\
\textbf{(ADVICE)} & \textbf{1} & \textbf{2} & \textbf{3} & \textbf{4} & \textbf{5} & \textbf{6} & \textbf{7} & \textbf{8} & \textbf{9} & \textbf{10} \\
\midrule
0 (Default) &<start\_of\_turn>& <bos>& user& \_only& Provide& Confidence& Answer& \textbackslash n & $```$ & Answer \\
100 & user & <start\_of\_turn> & Provide & \_only & \textbackslash n & \textbackslash n & <bos> & Confidence & Answer &  Answer \\
200 & <start\_of\_turn>& <bos>& user& Provide&\_only&\textbackslash n& \_A&  \textbf{\_Exile}& Confidence & \_between \\
300 & <start\_of\_turn>& \textbf{\_Exile}& user& <bos>& \_only& Provide& \_Kiss& Confidence& \_You& Question \\
400 & \textbf{\_Exile}& <start\_of\_turn>& \_only& <bos>& Provide& user& \_A& Confidence& Question & $```$ \\
500 & <start\_of\_turn>& <bos>& \textbf{\_Exile}& \_only& user& Provide& \_A& Confidence& $```$& Question \\
\bottomrule
\end{tabular}
\caption{Top 10 tokens sorted by their absolute attribution scores for \textsc{Gemma2-9b-it}. We observe an increase in the rank of the answer token (\textbf{\texttt{\_Exile}}), suggesting that ADVICE promotes greater answer dependence in the model.}
\label{tab:answer_groundedness_ig}
\end{table*}

\begin{figure}[t]
\centering
\begin{subfigure}{0.5\columnwidth}
\includegraphics[width=\columnwidth]{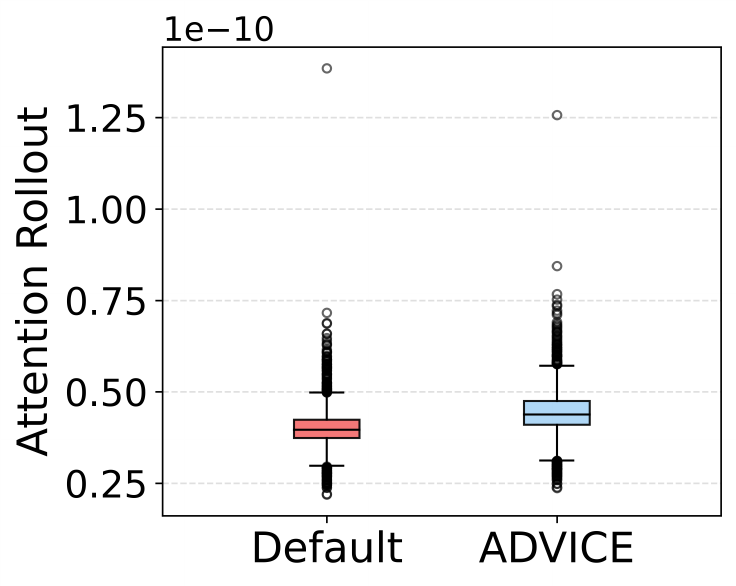}
\caption{\textsc{Gemma2-9b-it}}
\end{subfigure}\hfill
\begin{subfigure}{0.5\columnwidth}
\includegraphics[width=\columnwidth]{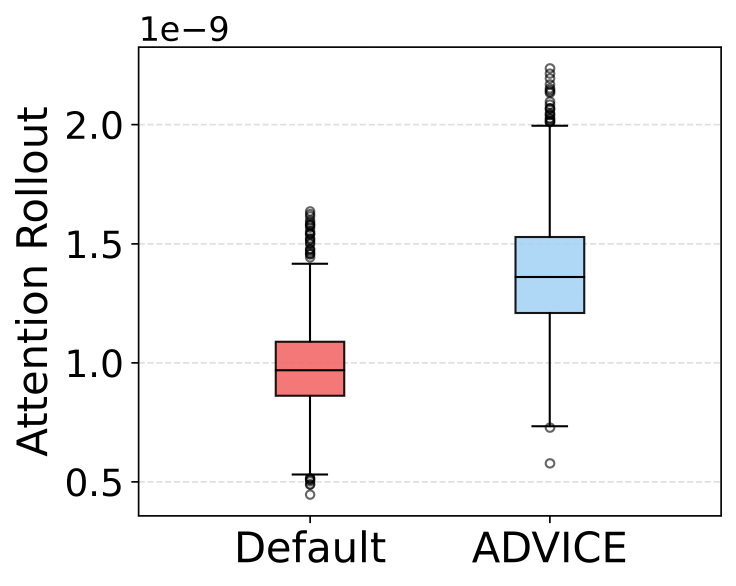}
\caption{\textsc{Llama3.1-8B-Instruct}}
\end{subfigure}
\caption{Attention Rollout score distributions for Confidence ($C$) $\rightarrow$ Answer ($A$), comparing ADVICE and Default. ADVICE contributes to improved attention. In both cases, the t-test confirms statistical significance.}
\label{fig:advice_vs_default}
\end{figure}

Second, we revisit the Attention Rollout analysis (\S\ref{sec:attribution}) to examine how adopting ADVICE alters the attention dynamics compared to the Default method (Default vs. ADVICE).
Figure \ref{fig:advice_vs_default} illustrates that compared to Default, ADVICE consistently directs the model’s attention more strongly toward the answer.
These results support our hypothesis that poor confidence verbalization arises from answer independence, and that ADVICE improves performance by mitigating this limitation.

Finally, we conduct a qualitative analysis of token attribution scores using Integrated Gradients, following the same procedure as in \S\ref{sec:attribution}.
Using a fixed input (\textit{Instruction}, $Q$, $A$), we track how token-level attribution patterns evolve throughout the fine-tuning process of ADVICE.
Specifically, we focus on the top-$k$ tokens ($k=10$) ranked by the absolute magnitude of their attribution scores, capturing both positive and negative contributions.
In Table \ref{tab:answer_groundedness_ig}, we can see that as training progresses, the rank of the answer token (\texttt{\_Exile}) increases, suggesting that ADVICE encourages the model to become more answer-dependent.

\begin{figure}[t]
\centering
\includegraphics[width=\columnwidth]{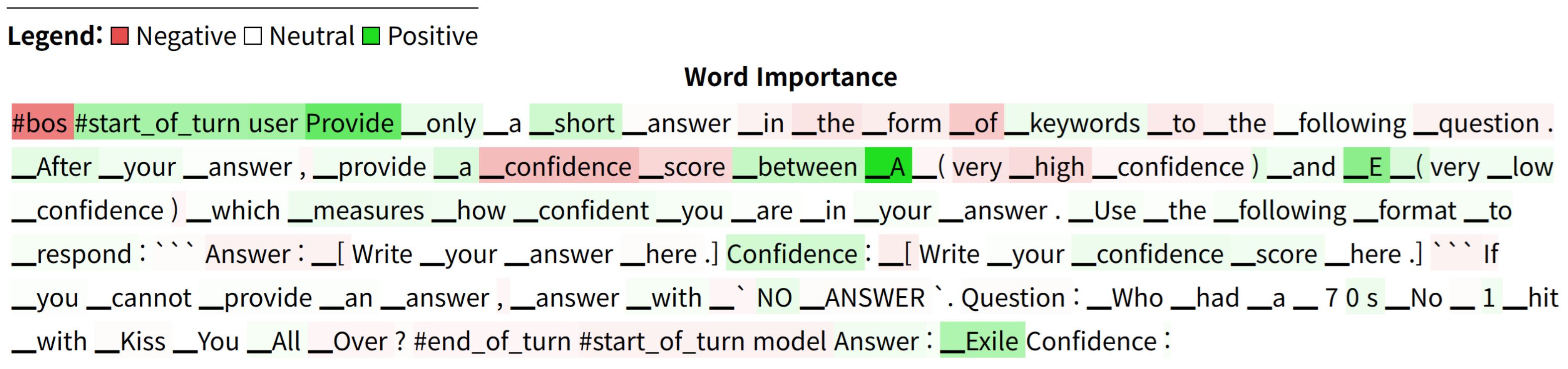}
\caption{Visualization of token attribution with Integrated Gradients (\textbf{ADVICE}, \textsc{Gemma2-9b-it}).}
\label{fig:integrated_gradients_advice}
\end{figure}

To enable a direct comparison with the Default, we contrast the Integrated Gradients visualizations before (Figure \ref{fig:integrated_gradients_default}) and after (Figure \ref{fig:integrated_gradients_advice}) applying ADVICE.
Under the Default setting, answer tokens are assigned negligible attribution scores, overshadowed by special and instruction tokens.
After training, the attribution assigned to answer tokens grows substantially, aligning with our quantitative results and confirming that ADVICE grounds confidence estimation in the generated answer.

In sum, our three experiments in this section consistently demonstrate that LLMs' overconfidence mainly arises from neglecting answer information in verbalized confidence estimation, and that ADVICE effectively mitigates this problem, by solving the primary cause of overconfidence in LLMs.

\section{Conclusion}
This work provides a systematic investigation into the fundamental cause of overconfidence in LLMs’ verbalized confidence.
In particular, our mathematical analysis identifies \textit{answer-independence} as the key contributing factor.
Based on this insight, we propose \textbf{ADVICE} (\textbf{A}nswer-\textbf{D}ependent \textbf{V}erbal\textbf{I}zed \textbf{C}onfidence \textbf{E}stimation), an intuitive and effective training framework that guides LLMs to generate more answer-grounded confidence estimation.

Extensive experiments show that ADVICE substantially mitigates the overconfidence commonly observed in LLMs, enabling them to produce more reliable and better-calibrated confidence estimates.
Furthermore, ADVICE generalizes to unseen verbalization formats and provides orthogonal gains when combined with existing methods, all with minimal token overhead. Finally, our post-hoc analyses confirm that these improvements are causally driven by enhanced answer dependence, validating answer independence as the diagnosis of overconfidence and its resolution as an effective remedy.

\section*{Limitations}

This study identifies the primary cause of overconfidence in LLMs and presents ADVICE, which effectively addresses it, leading to notable improvements in calibration.
However, several limitations remain, offering directions for future research.

First, while ADVICE enhances calibration through a contrastive objective that promotes answer-dependent confidence, it requires LLM-generated answers to form contrastive pairs, introducing additional data construction costs. 
Nevertheless, we consider this trade-off reasonable, as it explicitly targets the fundamental factor behind overconfidence and advances the development of more reliable models.

Second, this work primarily focuses on short-form QA and multiple-choice question answering. 
Extending the approach to tasks that demand long-context understanding and complex reasoning would be a valuable next step.

Third, calibration performance is inherently coupled with task accuracy: in high-accuracy regimes, even the Default method can appear well calibrated. 
For example, on SciQ---where models achieve over 90\% accuracy---the base model attains the best calibration performance (see Table \ref{tab:main_table_sciq} in the Appendix). 
We observe a similar pattern across other confidence estimation methods, highlighting the need for rigorous evaluation practices in the literature.

\section*{Acknowledgments}
This work was supported by Institute of Information \& communications Technology Planning \& Evaluation (IITP) grant funded by the Korea government(MSIT) (No.RS-2020-II201373, Artificial Intelligence Graduate School Program(Hanyang University)),  Institute of Information \& communications Technology Planning \& Evaluation (IITP) under the artificial intelligence semiconductor support program to nurture the best talents (IITP-(2026)-RS-2023-00253914) grant funded by the Korea government(MSIT), and the National Research Foundation of Korea(NRF) grant funded by the Korea government(MSIT) (RS-2025-00558151).

\bibliography{custom}

\clearpage

\appendix

\section{Experimental Setting Description}
\label{appendix:experimental_setting}
Here we provide the detailed settings for the experiments described in \S \ref{sec:comparison_distributions} and \S \ref{sec:settings}.

First, by comparing confidence distributions in \S \ref{sec:comparison_distributions}, we further demonstrate the answer independence of verbalized confidence.
For evaluating answer-independence, we leverage the training set of TriviaQA.
We construct multiple answers corresponding to the same question in the dataset, i.e., $(q, \{a_1, \dots, a_m\})$.
We set $m=30$ for TriviaQA to reflect its free-form generation setting, whereas for MCQ datasets we set $m=4$ by leveraging the pre-defined distractors provided with each dataset.
We then remove duplicate answers to construct the $\hat{A}_q=\{a_1,\dots,a_n\}$ for each question in dataset.
Note that the number of filtered answers, $n$, depends on the question $q$.
Furthermore, it is important to emphasize that the answer candidates $a_i$ and $a_j$ are selected based on their distinct informational content, rather than their ground-truth correctness.
The objective of this analysis is to demonstrate that even when the model is presented with two different answers containing different information, the verbalized confidence remains nearly identical.

Second, we also provide detailed explanations for metrics used in \S \ref{sec:settings}.
ECE is defined as follows:
\begin{equation*}
\mathrm{ECE}=\sum_{m=1}^M  {\frac{|B_m|}{N}}|\mathrm{acc}(B_m)-\mathrm{conf}(B_m)|,
\end{equation*}
where $M$ denotes the number of bins, $N$ the total number of samples, $B_m$ the collection of instances assigned to the $m$-th bin, $\mathrm{acc}$ the accuracy, and $\mathrm{conf}$ the confidence.
We set $M=10$, a value commonly used in practice.
ECE quantifies the average absolute difference between predicted confidence and empirical accuracy over grouped confidence intervals.

We also employ NCE, a variation of ECE, to complement each other.
We modify NCE by taking its absolute value for more intuitive interpretation, so that smaller value indicates better calibration.

|NCE| is formulated as:
\begin{equation*}
|\mathrm{NCE}|=|\sum_{m=1}^M  {\frac{|B_m|}{N}}(\mathrm{acc}(B_m)-\mathrm{conf}(B_m))|.
\end{equation*}
The distinction is that NCE computes a weighted sum of signed differences, whereas ECE computes one of absolute differences.
As a result, biased confidence estimation, such as over- or under-confidence, yields a large absolute NCE value.

The Brier score is defined as the mean squared difference between predicted confidence scores ($c_n$) and true binary outcomes ($y_n$), directly measuring the accuracy of probabilistic predictions.
It is calculated as:
\begin{equation*}
\mathrm{BS}=\frac{1}{N}\sum_{n=1}^N (y_n - c_n)^2.
\end{equation*}

Finally, AUROC measures the likelihood that a randomly selected positive instance receives a higher confidence score than a randomly selected negative one, reflecting the model’s overall ability to rank predictions by confidence.

\section{LLM Probing Methods}
\label{sec:appendix_probing_methods}
\paragraph{Attention Rollout}
Compared to naive attention scores, Attention Rollout provides more reliable attributions by recursively aggregating attention across layers. This aggregation accounts for the residual connections and the hierarchical flow of information, yielding a more faithful estimate of token contributions.

Attention Rollout is recursively defined as:
\begin{equation*}
\tilde{A}(l_i) =
\begin{cases}
A(l_i)\tilde{A}(l_{i-1}), & \text{if } i > j, \\
A(l_i), & \text{if } i = j,
\end{cases}
\end{equation*}
where $A(l_i)$ denotes the raw attention matrix of layer $i$, updated with residual connections and computed as 
\[
A(l_i) = 0.5\,W_{\text{att}, i} + 0.5I.
\]
We define the Question tokens as those ranging from \texttt{Question:} to the end of the input (i.e., the \texttt{<end\_of\_turn>} token), and the Answer tokens as those spanning from \texttt{Answer:} to the token immediately preceding the subsequent \texttt{Confidence:}.
As the next step, we computed the attention rollout for each token, starting from the position of the colon (i.e., the ``\texttt{:}'' in \texttt{Answer:}), which corresponds to the point where the first token of the answer or the confidence expression begins to be generated.
Subsequently, we compute the rollout scores across the entire layer, and aggregate the rollout scores by taking their average.

\paragraph{Integrated Gradients}
Integrated Gradients are formulated as:
\begin{equation*}
(x_i - x_i')\times\int_{\alpha=0}^{1} {\frac{\partial F(x'+\alpha \times (x-x'))}{\partial x_i}} d \alpha,e
\end{equation*}
where $i$ denotes the feature dimension, and $x_i'$ corresponds to the baseline input. 
In practice, the integral is approximated via a Riemann sum with a predefined number of interpolation steps, $n\_steps$.
We employed the \texttt{IntegratedGradients} implementation from the \texttt{captum} library to compute attribution scores. 
For all experiments, we set the hyperparameters to $n\_steps=1024$ and $internal\_batch\_size=32$, and adopted a zero vector as the baseline. 
Furthermore, we visualized the resulting attributions using the visualization utilities provided within the same package.

\section{Implementation Details}
\label{appendix:implementation_details}

\paragraph{Training Details}
Here, we describe the implementation details of ADVICE.
We utilize LoRA \cite{hu2022lora} from the HuggingFace PEFT library \cite{peft} for fine-tuning. 

Specifically, we fine-tune the adapters attached to the query, key, value, and output projection modules across all transformer layers, using a rank of $r=16$ and a scaling factor of $\alpha=32$.
Optimization is performed with AdamW at a learning rate $\eta$, learning rate warmup over the first 5\% steps, and linear decay of the learning rate ($\eta=3 \times 10^{-5}$ for \textsc{Mistral-7B-Instruct-v0.3}, $\eta=1 \times 10^{-5}$ for the rest).
We adopt a batch size of 16 and apply gradient accumulation with a factor of 2 using the Accelerate framework.
We train all LLMs for 4 epochs.
All training runs are conducted on 1 NVIDIA H200 NVL PCIe GPU.

\begin{figure}[t]
\centering
\includegraphics[width=0.7\columnwidth]{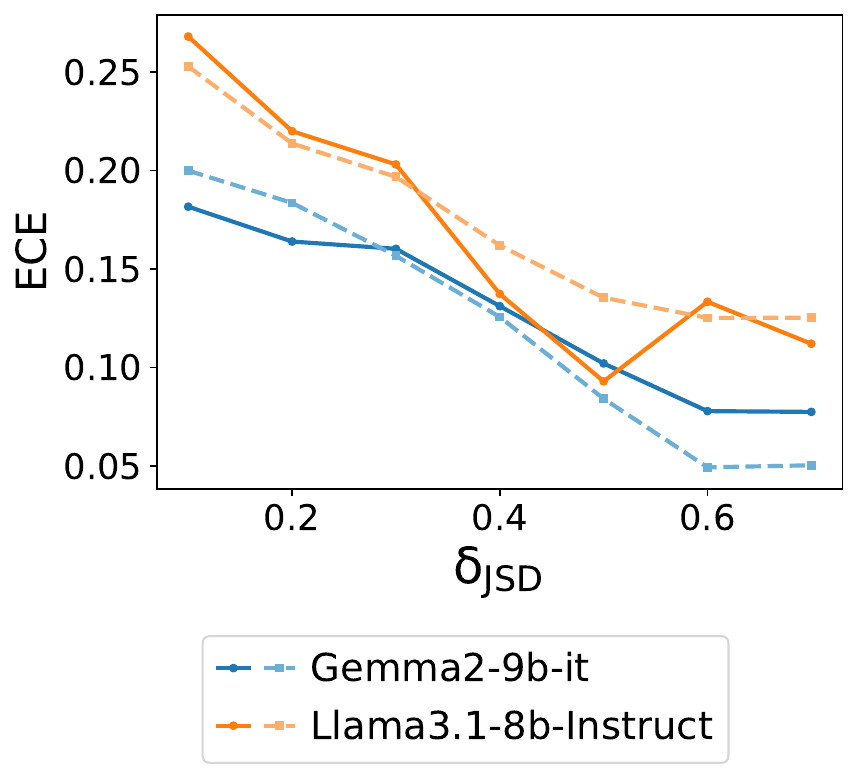}
\caption{ECE as a function of $\delta_{\mathrm{JSD}}$ on TriviaQA. \textcolor{blue}{Blue} lines correspond to \textsc{Gemma-2-9b-it}, and \textcolor{orange}{orange} lines to \textsc{LLaMA-3.1-8B-Instruct}.}
\label{fig:hyperparameter}
\end{figure}

Based on the results in Figure \ref{fig:hyperparameter}, we set $\delta_{\mathrm{JSD}}$ to $0.6$, as the Jensen–Shannon divergence (JSD) is defined to take values between $0$ and $\ln{2}\ (\approx 0.693)$.
The value of $\delta_{\mathrm{Margin}}$ is set to $1$, which is chosen to be strictly greater than the expectation difference observed in all experimental settings.
Explanation for each training objective hyperparameter is described in \S \ref{sec:training_objective}.

Note that the size of our training dataset varies depending on the generated texts of each LLM, as our dataset construction process leverages the model itself to generate samples using stochastic decoding.
Consequently, we obtain nearly 1k, 1k, and 2k training samples for \textsc{Gemma-2-9b-it}, \textsc{Llama-3.1-8B-Instruct}, and \textsc{Mistral-7B-Instruct-v0.3}, respectively.

For optimization stability, we impose within-batch homogeneity: although training employs multiple verbalization variants, each mini-batch contains a single verbalization type, and batches are shuffled across steps.

\paragraph{Self-Consistency}
Following \citet{xiong2024can}, we implement the method using the vanilla prompt with $M=5$.
Specifically, we select the Avg-Conf variant of the method, which computes the weighted sum of confidence scores and this configuration has been shown to outperform other ones.
This involves prompting the LLM to generate five candidate answers and aggregating them as follows:
\begin{equation*}
C_{\text{conf}} = 
\frac{
\sum_{i=1}^{M} \mathcal{I}\{\hat{Y}_i = \tilde{Y}\} \times C_i
}{
\sum_{i=1}^{M} C_i
},
\end{equation*}
where $\hat{Y}_i$ are candidate answers with their corresponding verbalized confidence $C_i$ and $\mathcal{I}$ is indicator function.
Note that $\tilde{Y}$ denotes the answer that has the highest confidence score among all candidate answers.

\paragraph{ConfTuner}
We re-implement ConfTuner based on their official code.\footnote{\url{https://github.com/liushiliushi/ConfTuner}}
For \textsc{Llama-3.1-8B-Instruct}, we use their publicly available fine-tuned model.\footnote{\href{https://huggingface.co/liushiliushi/ConfTuner-LLaMA}{liushiliushi/ConfTuner-LLaMA}}
We fine-tune \textsc{Gemma-2-9b-it} and \textsc{Mistral-7B-Instruct-v0.3} on our training dataset.
Following the original implementation, we adopt the same prompt type (i.e., \textbf{ScoreNumber}).
Since the number of training samples differs from the original paper, we also adjust the number of training epochs accordingly: we fine-tune \textsc{Mistral-7B-Instruct-v0.3} for 3 epochs and \textsc{Gemma-2-9b-it} for 2 epochs.

\paragraph{ADVICE \textit{w/} ConfTuner}
We integrate our training objective with ConfTuner’s calibration loss.
Specifically, using the correct and incorrect answers in our training dataset, we adapt ConfTuner’s calibration loss to our setting and re-define $\mathcal{L}{\mathrm{cal}}$ accordingly.
We then train three models with five objectives, i.e., $\mathcal{L}_\mathrm{LM}, \mathcal{L}_\mathrm{JSD}, \mathcal{L}_\mathrm{Margin}, \mathcal{L}_\mathrm{Sum}, \mathcal{L}_\mathrm{cal}$, following the definition of $\mathcal{L}_\mathrm{cal}$ in \citet{li2025conftunertraininglargelanguage}. We set $\lambda_{\mathrm{LM}}$, $\lambda_{\mathrm{JSD}}$, $\lambda_{\mathrm{Margin}}$, and $\lambda_{\mathrm{Sum}}$ to 0.5, and all coefficients of $\mathcal{L}_\mathrm{cal}$ to 1.
We fine-tune \textsc{Gemma-2-9b-it} for 3 epochs and \textsc{Mistral-7B-Instruct-v0.3} for 4 epochs.
We optimize with AdamW at a learning rate $\eta = 3 \times 10^{-5}$.
Moreover, we further train their fine-tuned model using the loss described in \S \ref{sec:training_objective} for \textsc{Llama-3.1-8B-Instruct}.

\section{Confidence Verbalization Types}
\label{appendix:confidence_types}

As outlined in \S \ref{sec:settings}, we utilize five types of verbalization—\textbf{ScoreLetter}, \textbf{ScoreNumber}, \textbf{ScoreText}, \textbf{ScoreFloat} and \textbf{ScorePercent}.
We train ADVICE with \textbf{ScoreLetter} and \textbf{ScoreNumber} and evaluate on (1) the same two prompt types and (2) the other prompt types, where the latter serves as a generalization test in Table \ref{tab:prompt_generalization}.
Following \citet{li2025conftunertraininglargelanguage}, we train and evaluate LLMs in the \textbf{ScoreNumber} setting for ConfTuner.
To quantify calibration metrics, each verbalized confidence expression is mapped to a numeric value within the interval $[0, 1]$. We specify the numeric mappings for each prompt type as follows:
\begin{itemize}[itemsep=0pt, topsep=5pt, leftmargin=13pt]
    \item \textbf{ScoreLetter}: Each letter token \{\texttt{E}, \texttt{D}, \texttt{C}, \texttt{B}, \texttt{A}\} is mapped to: 
    \texttt{E}$=0.1$, \texttt{D}$=0.3$, \texttt{C}$=0.5$, \texttt{B}$=0.7$, \texttt{A}$=0.9$.
    \item \textbf{ScoreNumber}: Each digit $i \in \{\texttt{0},\texttt{1} \ldots, \texttt{9}\}$ is assigned a value of $i/9$.
    \item \textbf{ScoreText}: Verbalized levels are mapped as 
    \texttt{low} $ = 0.1 $, \texttt{medium} $= 0.5$, \texttt{high} $= 0.9$.
    \item \textbf{ScoreFloat}: Each floating-point value is used directly without further mapping.
    \item \textbf{ScorePercent}: Each percentage token $i\%$ is mapped to a value of $i/100$.
\end{itemize}

\section{Prompt Templates}
We provide the prompt templates, as shown in Table \ref{tab:prefix_prompt} and Table \ref{tab:main_prompts}, following the formats used by \citet{yang2025on}.
For ConfTuner, we use the template in Table \ref{tab:conftuner_prompt} proposed by \citet{li2025conftunertraininglargelanguage}.

\begin{table}[t]
\scriptsize
\setlength{\tabcolsep}{0.5pt}
\renewcommand{\arraystretch}{1.1}
\begin{tabular}{llllll}
\toprule
\multirow{2}[2]{*}{\textbf{Model}} & \multirow{2}[2]{*}{\textbf{Method}} & \multicolumn{4}{c}{\textbf{SciQ}} \\ 
\cmidrule{3-6}
& & \phantom{0}ECE & \phantom{0}|NCE| & \phantom{00}BS & AUROC \\ \midrule
\multirow{5}{*}{\shortstack[l]{\textsc{Gemma2} \\ \textsc{9b-it}}} & Default & \phantom{00}4.9 & \phantom{0}\phantom{0}1.5 & \phantom{0}\phantom{0}\textbf{5.2} & \phantom{0}50.3 \\ 
& Prompting & \phantom{00}5.1 & \phantom{00}\textbf{1.4} & \phantom{00}\textbf{5.2} & \phantom{0}50.0 \\
& Self-Consistency & \phantom{00}\textbf{4.0}$_{\pm 0.1}$ & \phantom{00}4.0$_{\pm 0.1}$ & \phantom{00}6.3$_{\pm 0.1}$ & \phantom{0}70.3$_{\pm 0.6}$ \\ 
& ConfTuner & \phantom{00}6.8$_{\pm 0.7}$ & \phantom{00}2.9$_{\pm 0.8}$ & \phantom{00}6.9$_{\pm 0.4}$ & \phantom{0}\textbf{77.2}$_{\pm 0.5}$ \\ 
& \textbf{ADVICE (Ours)} & \phantom{0}14.2$_{\pm 4.4}$ & \phantom{0}14.2$_{\pm 4.4}$ & \phantom{00}7.4$_{\pm 1.5}$ & \phantom{0}72.6$_{\pm 2.8}$ \\ \midrule
\multirow{5}{*}{\shortstack[l]{\textsc{Llama3.1} \\ \textsc{8B Instruct}}} & Default & \phantom{00}\textbf{3.7} & \phantom{00}2.1 & \phantom{00}8.2 & \phantom{0}52.0 \\ 
& Prompting & \phantom{00}4.3 & \phantom{00}\textbf{0.7} & \phantom{00}7.2 & \phantom{0}51.4 \\
& Self-Consistency & \phantom{00}\textbf{1.5}$_{\pm 0.3}$ & \phantom{00}1.3$_{\pm 0.2}$ & \phantom{00}7.1$_{\pm 0.2}$ & \phantom{0}\textbf{74.0}$_{\pm 1.0}$ \\ 
& ConfTuner & \phantom{00}9.5 & \phantom{00}9.1 & \phantom{00}9.5 & \phantom{0}58.4 \\ 
& \textbf{ADVICE (Ours)} & \phantom{00}7.9$_{\pm 2.1}$ & \phantom{00}7.7$_{\pm 2.2}$& \phantom{00}\textbf{6.6}$_{\pm 0.3}$ & \phantom{0}73.4$_{\pm 3.5}$ \\ \midrule
\multirow{5}{*}{\shortstack[l]{\textsc{Mistral} \\ \textsc{7B Instruct}}} & Default & \phantom{0}10.3 & \phantom{00}9.0 & \phantom{0}15.3 & \phantom{0}50.3 \\ 
& Prompting & \phantom{00}9.8 & \phantom{00}8.5 & \phantom{0}14.9 & \phantom{0}51.7 \\
& Self-Consistency & \phantom{0}14.8$_{\pm 0.4}$ & \phantom{0}14.5$_{\pm 0.3}$ & \phantom{0}16.9$_{\pm 0.4}$ & \phantom{0}67.1$_{\pm 0.7}$ \\ 
& ConfTuner & \phantom{0}10.8$_{\pm 1.1}$ & \phantom{00}9.9$_{\pm 1.7}$ & \phantom{0}14.0$_{\pm 0.5}$ & \phantom{0}\textbf{68.6}$_{\pm 0.9}$ \\ 
& \textbf{ADVICE (Ours)} & \phantom{00}\textbf{8.3}$_{\pm 0.9}$ & \phantom{00}\textbf{8.1}$_{\pm 1.1}$ & \phantom{0}\textbf{12.5}$_{\pm 0.7}$ & \phantom{0}63.0$_{\pm 3.8}$ \\ 
\bottomrule
\end{tabular}
\caption{Average performance over trained verbalization types (i.e., Score\{Letter, Number\} for ADVICE), evaluated on SciQ. Values are percentages. Best results are in \textbf{bold}—minimum for ECE and BS, absolute minimum for NCE, and maximum for AUROC.}
\label{tab:main_table_sciq}
\end{table}
\section{Effect of Hyperparameter}
\label{sec:hyperparmeter}
To examine the impact of $\mathcal{L}_{\mathrm{JSD}}$, we evaluate the calibration performance under varying $\delta_{\mathrm{JSD}}$.
The hyperparameter $\delta_{\mathrm{JSD}}$ controls how sensitively the model distinguishes between the two answer distributions, $P_{\text{correct}}$ and $P_{\text{wrong}}$.
Figure \ref{fig:hyperparameter} shows the variation of ECE across different values of $\delta_{\mathrm{JSD}}$.
We consistently observe a reduction in ECE as $\delta_{\mathrm{JSD}}$ increases. 
This is intuitive, as a smaller $\delta_{\mathrm{JSD}}$ reduces the penalty for similarity between contrastive distributions, resulting in less distinct separation and degraded calibration performance.

\section{Qualitative Evaluation}
\begin{figure*}[t]
\centering
\begin{subfigure}{0.85\textwidth}
\includegraphics[width=\textwidth]{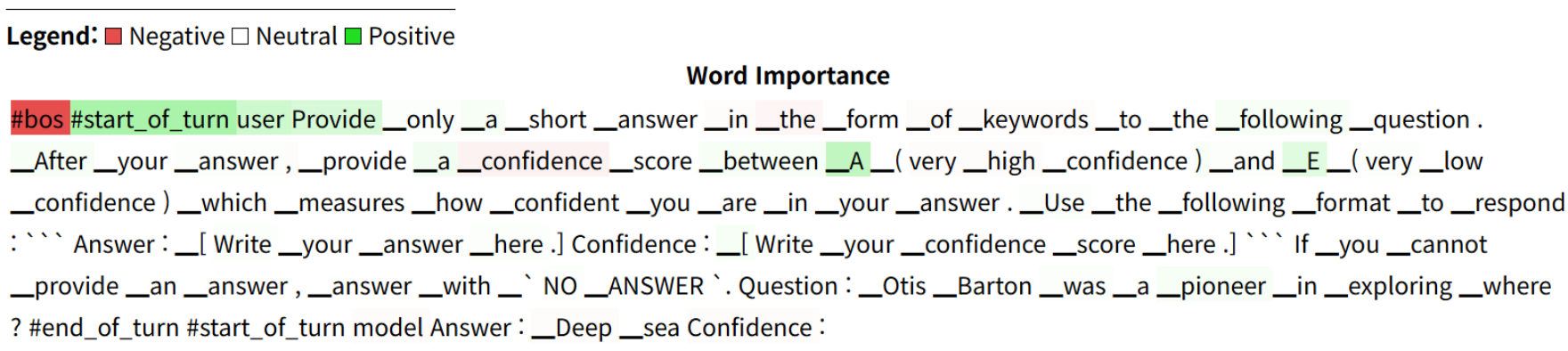}
\caption{Default}
\end{subfigure}
\begin{subfigure}{0.85\textwidth}
\includegraphics[width=\textwidth]{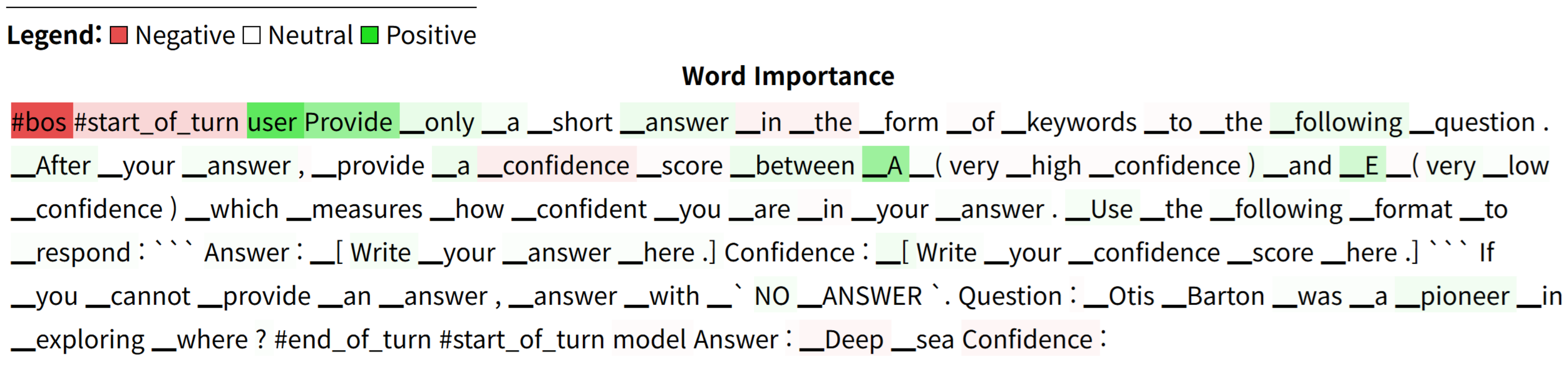}
\caption{100 Step}
\end{subfigure}
\begin{subfigure}{0.85\textwidth}
\includegraphics[width=\textwidth]{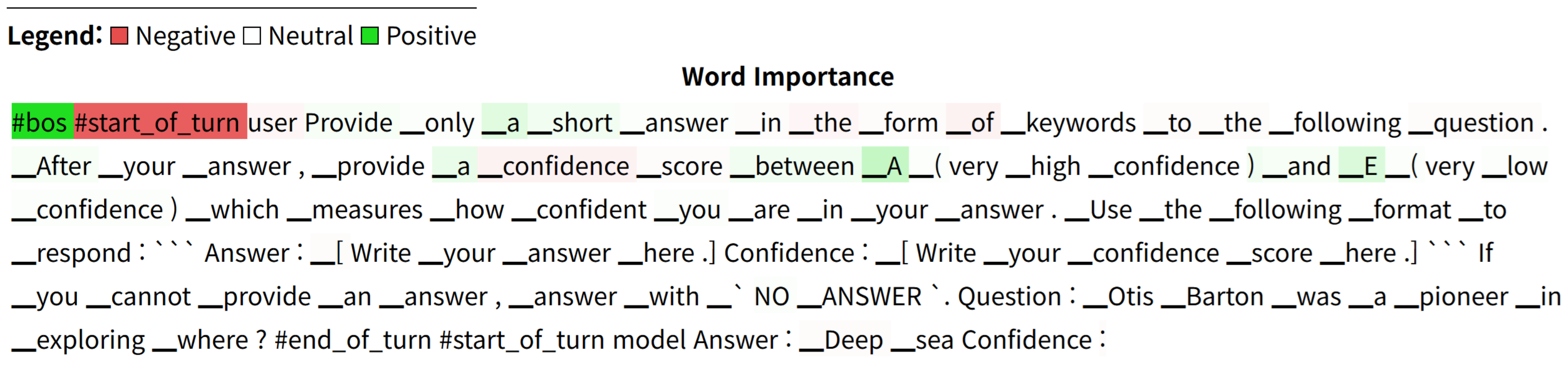}
\caption{200 Step}
\end{subfigure}
\begin{subfigure}{0.85\textwidth}
\includegraphics[width=\textwidth]{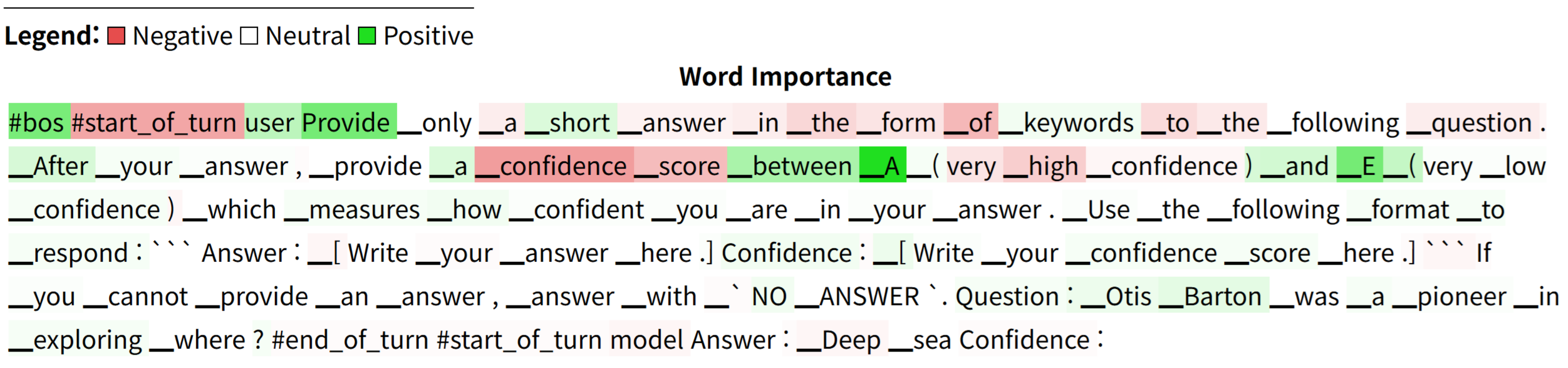}
\caption{300 Step}
\end{subfigure}
\begin{subfigure}{0.85\textwidth}
\includegraphics[width=\textwidth]{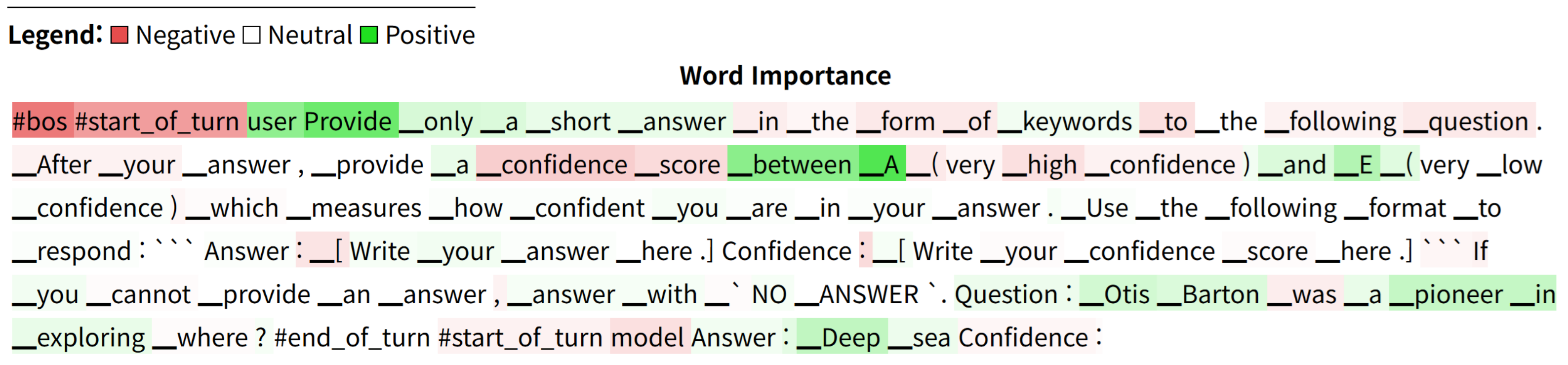}
\caption{400 Step}
\end{subfigure}
\caption{
Visualization of token attribution changes across training steps using Integrated Gradients (\textsc{Gemma2-9b-it}).
As training progresses, the attribution scores on answer tokens consistently increase.
}
\label{fig:step_ablation_ig_gemma}
\end{figure*}

\begin{figure*}[t]
\centering
\begin{subfigure}{0.75\textwidth}
\includegraphics[width=\textwidth]{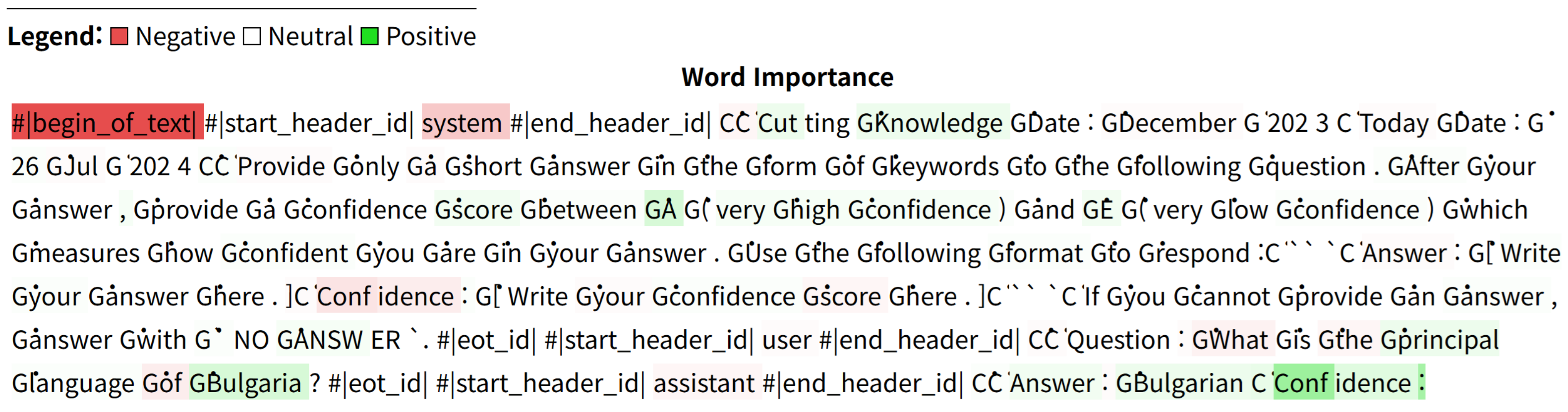}
\caption{Default}
\end{subfigure}
\begin{subfigure}{0.75\textwidth}
\includegraphics[width=\textwidth]{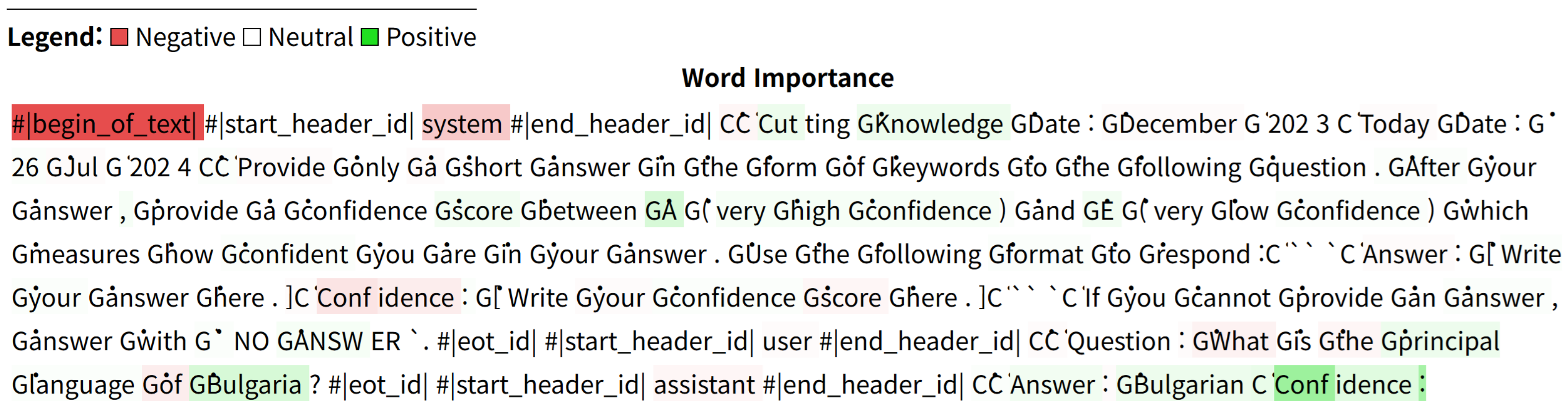}
\caption{100 Step}
\end{subfigure}
\begin{subfigure}{0.75\textwidth}
\includegraphics[width=\textwidth]{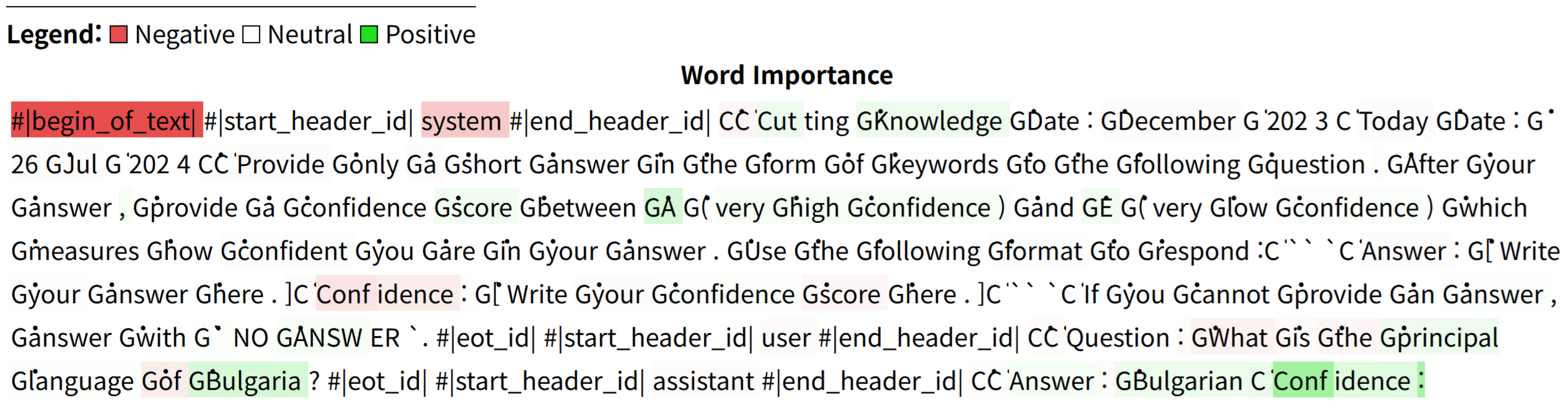}
\caption{200 Step}
\end{subfigure}
\begin{subfigure}{0.75\textwidth}
\includegraphics[width=\textwidth]{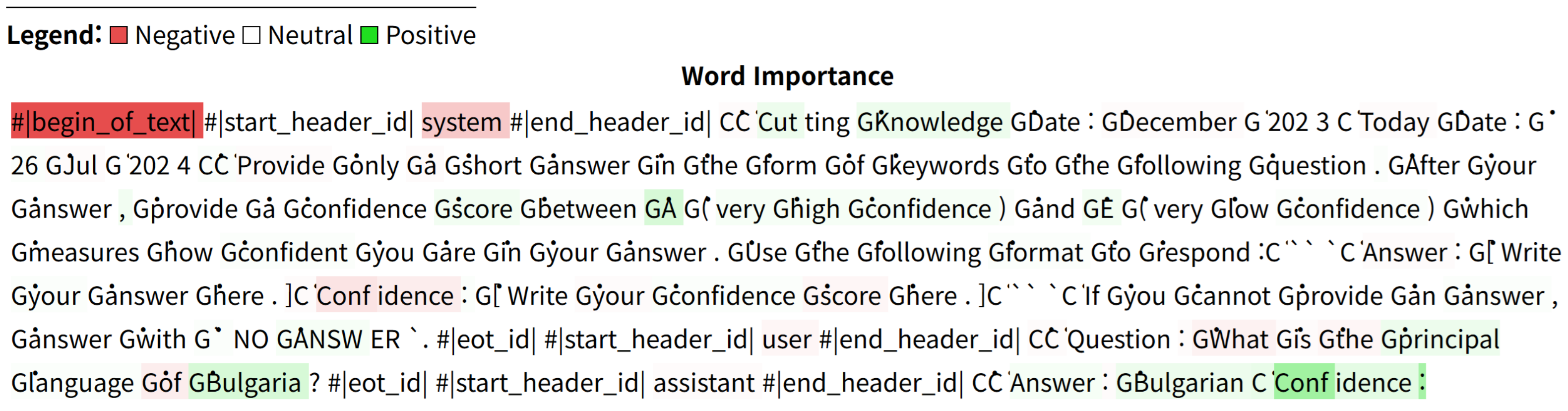}
\caption{300 Step}
\end{subfigure}
\begin{subfigure}{0.75\textwidth}
\includegraphics[width=\textwidth]{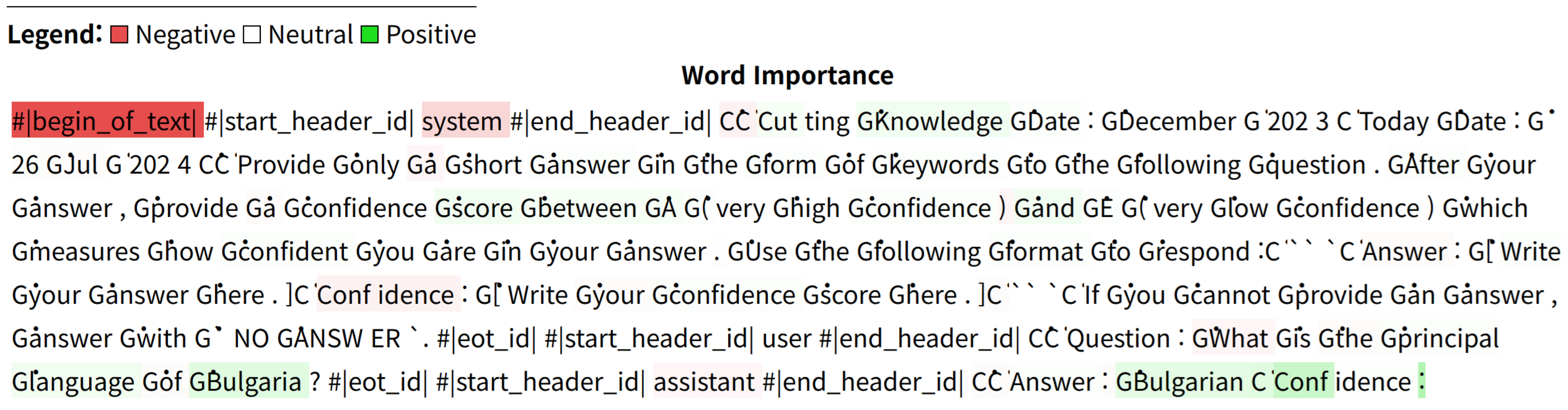}
\caption{400 Step}
\end{subfigure}
\begin{subfigure}{0.75\textwidth}
\includegraphics[width=\textwidth]{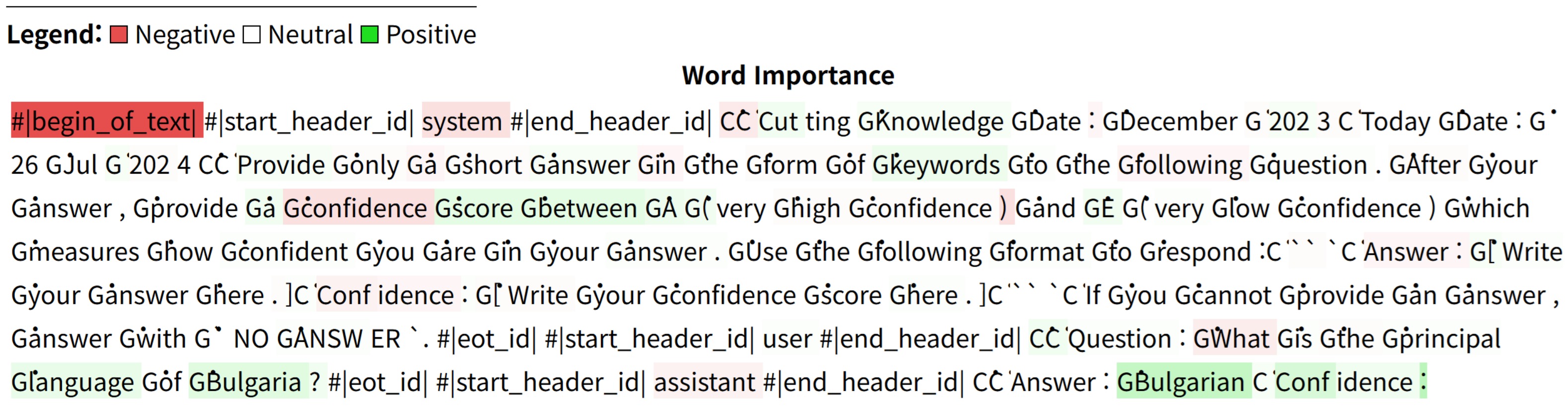}
\caption{500 Step}
\end{subfigure}
\caption{
Visualization of token attribution changes across training steps using Integrated Gradients (\textsc{Llama3.1-8B-Instruct}).
As training progresses, the attribution scores are reallocated.
}
\label{fig:step_ablation_ig_llama}
\end{figure*}

We qualitatively assess how our method affects the extent to which confidence is grounded in the answer.
In Figure \ref{fig:step_ablation_ig_gemma} and Figure \ref{fig:step_ablation_ig_llama}, we observe that as training progresses, the attribution scores of answer tokens gradually increase.
This result demonstrates that our method enhances calibration capability by inducing answer-dependent confidence estimation.

\begin{figure*}[t]
\centering
\begin{subfigure}{0.4\textwidth}
\includegraphics[width=\textwidth]{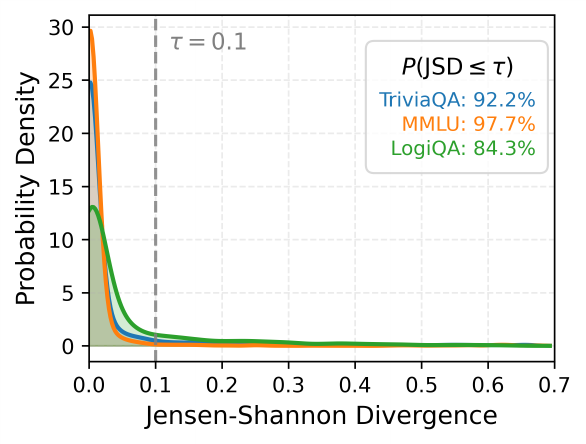}
\caption{\textsc{Gemma2-9b-it}}
\end{subfigure}
\begin{subfigure}{0.4\textwidth}
\includegraphics[width=\textwidth]{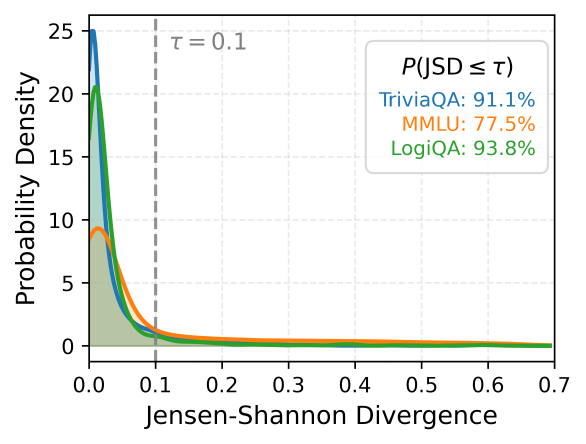}
\caption{\textsc{Llama3.1-8B-Instruct}}
\end{subfigure}
\caption{Probability density functions (PDFs) of the set $\{\mathrm{JSD}(P_M(C | q, a_i) || P_M(C | q,a_j))\}$, where $M \in \{\textsc{Gemma2}, \textsc{LLaMA3.1}\}$ and $(q,a_i,a_j)$ are from TriviaQA, MMLU, LogiQA under ScoreNumber setting.
Each PDF (solid curve) is computed via Gaussian kernel density estimation.
Near-zero concentration implies answer-independent confidence.}
\label{fig:answer_independence_scorenumber}
\end{figure*}

\begin{figure*}[t]
\centering
\begin{subfigure}{0.4\textwidth}
\includegraphics[width=\textwidth]{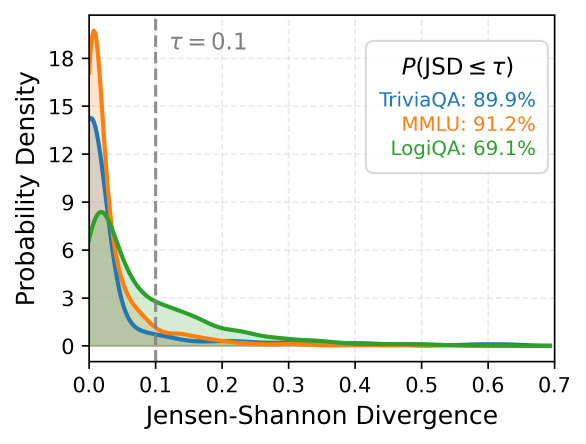}
\caption{\textsc{Gemma2-9b-it}}
\end{subfigure}
\begin{subfigure}{0.4\textwidth}
\includegraphics[width=\textwidth]{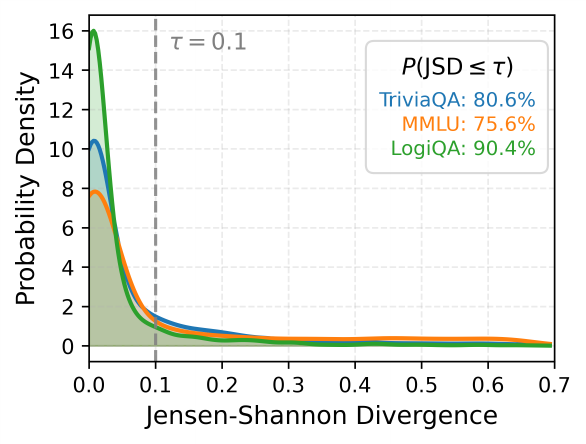}
\caption{\textsc{Llama3.1-8B-Instruct}}
\end{subfigure}
\caption{Probability density functions (PDFs) of the set $\{\mathrm{JSD}(P_M(C | q, a_i) || P_M(C | q,a_j))\}$, where $M \in \{\textsc{Gemma2}, \textsc{LLaMA3.1}\}$ and $(q,a_i,a_j)$ are from TriviaQA, MMLU, LogiQA under ScoreLetter setting.
In this figure, $a_i$ denotes the correct answer for $q$, while $a_j$ is restricted to one of the sampled wrong answers; all other settings are identical to Figure \ref{fig:answer_independence}.
Consistent with Figure \ref{fig:answer_independence}, this results reveal little change in the confidence distribution between correct and wrong answers, suggesting that verbalized confidence is largely insensitive to answer correctness.
}
\label{fig:answer_independence_scoreletter_correct_wrong}
\end{figure*}

\begin{figure*}[t]
\centering
\begin{subfigure}{0.38\textwidth}
\includegraphics[width=\textwidth]{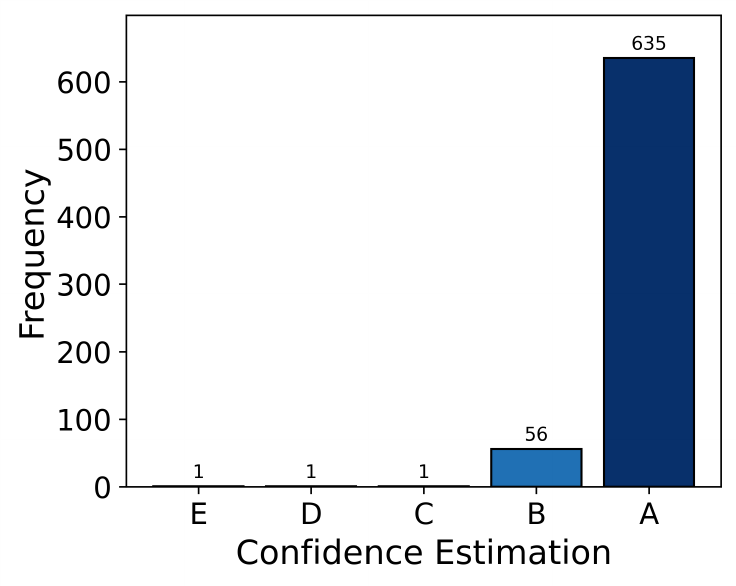}
\caption{Correct answer}
\end{subfigure}
\begin{subfigure}{0.38\textwidth}
\includegraphics[width=\textwidth]{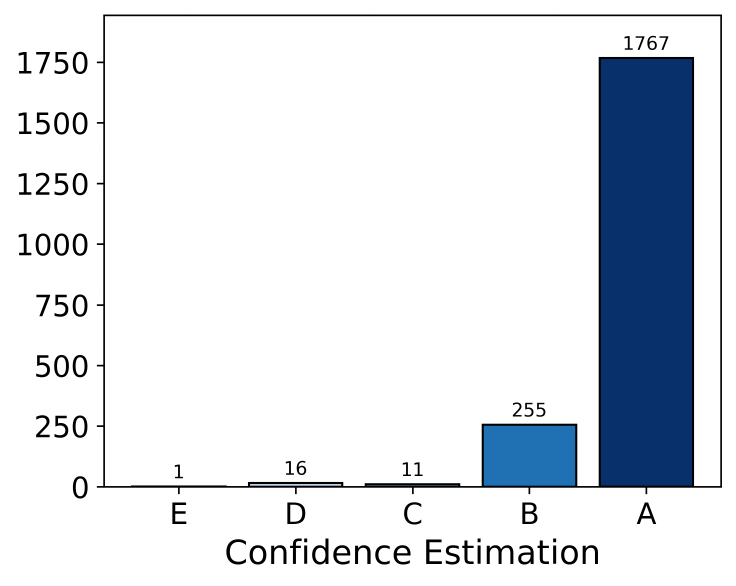}
\caption{Wrong answer}
\end{subfigure}
\caption{Histograms of verbalized confidence for correct and wrong answers. Both distributions are skewed toward high confidence, indicating that LLMs often express high confidence regardless of correctness.
Together with Figure \ref{fig:answer_independence}, this motivates comparing different incorrect answers to assess whether verbalized confidence is answer-dependent.
}
\label{fig:correct_wrong_confidence_estimation}
\end{figure*}

\clearpage

\begin{figure*}[t]
\centering
\begin{subfigure}{0.23\textwidth}
  \centering
  \includegraphics[width=\linewidth]{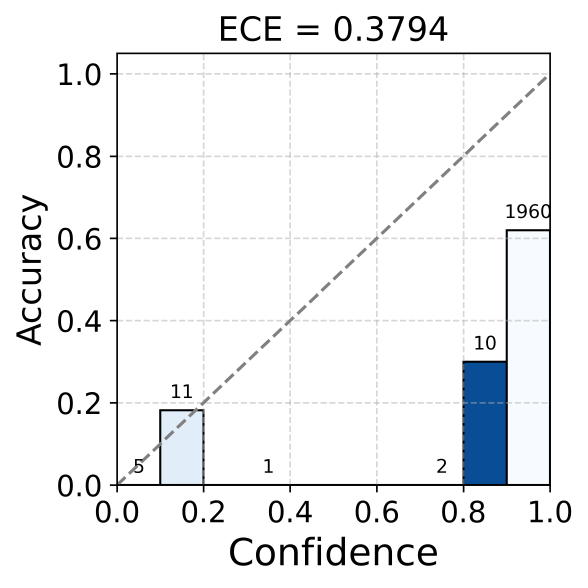}
    \caption{Default}
\end{subfigure}
\begin{subfigure}{0.23\textwidth}
  \centering
  \includegraphics[width=\linewidth]{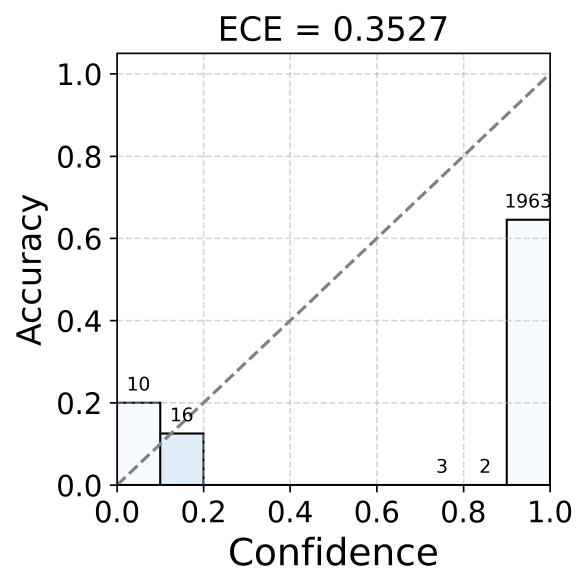}
    \caption{Prompting}
\end{subfigure}
\begin{subfigure}{0.23\textwidth}
  \centering
  \includegraphics[width=\linewidth]{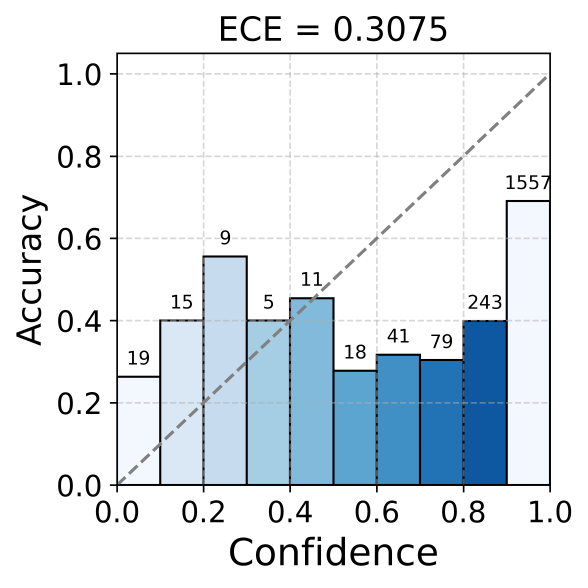}
    \caption{Self-Consistency}
\end{subfigure}
\par
\begin{subfigure}{0.23\textwidth}
  \centering
  \includegraphics[width=\linewidth]{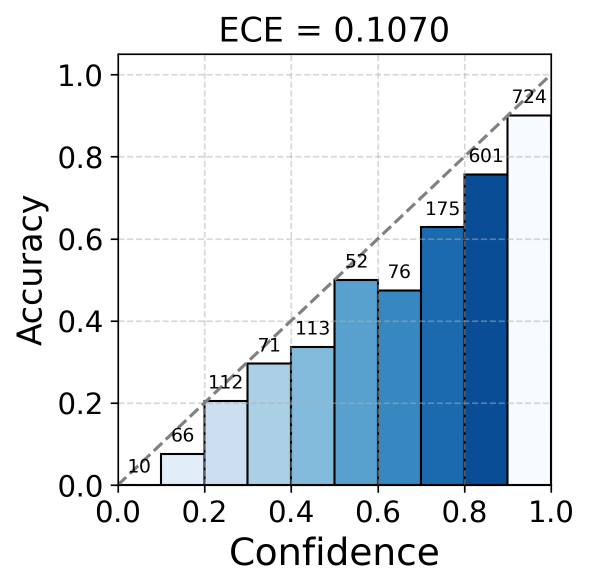}
    \caption{ConfTuner}
\end{subfigure}
\begin{subfigure}{0.23\textwidth}
  \centering
  \includegraphics[width=\linewidth]{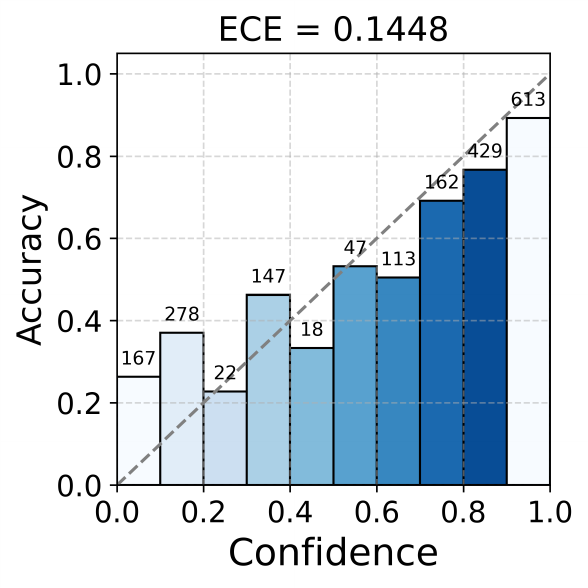}
    \caption{ADVICE}
\end{subfigure}
\begin{subfigure}{0.23\textwidth}
  \centering
  \includegraphics[width=\linewidth]{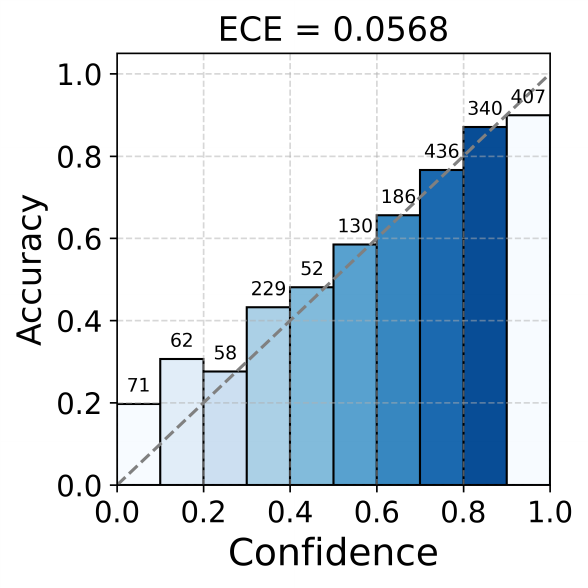}
    \caption{ADVICE \textit{w/} ConfTuner}
\end{subfigure} 
\caption{Reliability diagrams of \textsc{Mistral-7B-Instruct-v0.3} on TriviaQA under the ScoreNumber setting, where numbers above each bin indicate the number of instances.}
\label{fig:reliability_diagram_appendix1}
\end{figure*}

\begin{figure*}[t]
\centering
\begin{subfigure}{0.23\textwidth}
  \centering
  \includegraphics[width=\linewidth]{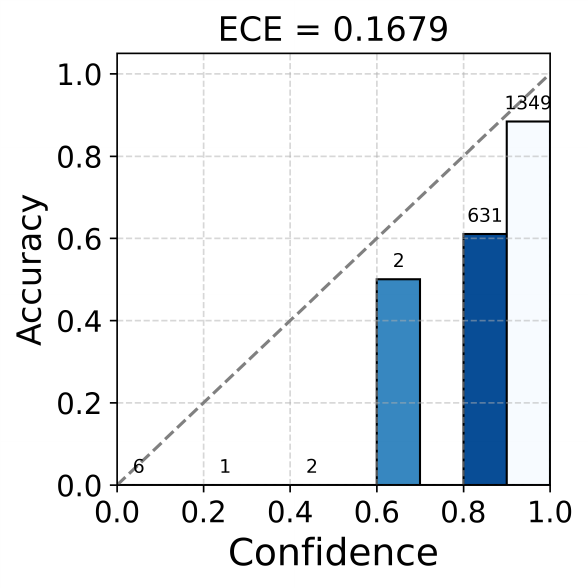}
    \caption{Default}
\end{subfigure}
\begin{subfigure}{0.23\textwidth}
  \centering
  \includegraphics[width=\linewidth]{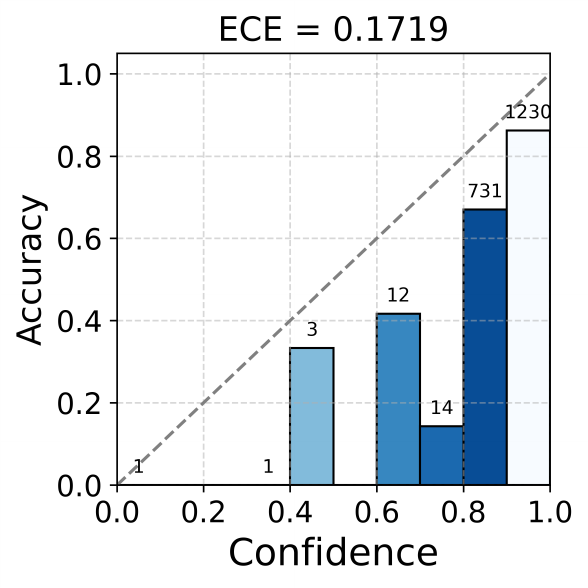}
  \caption{Prompting}
\end{subfigure}
\begin{subfigure}{0.23\textwidth}
  \centering
  \includegraphics[width=\linewidth]{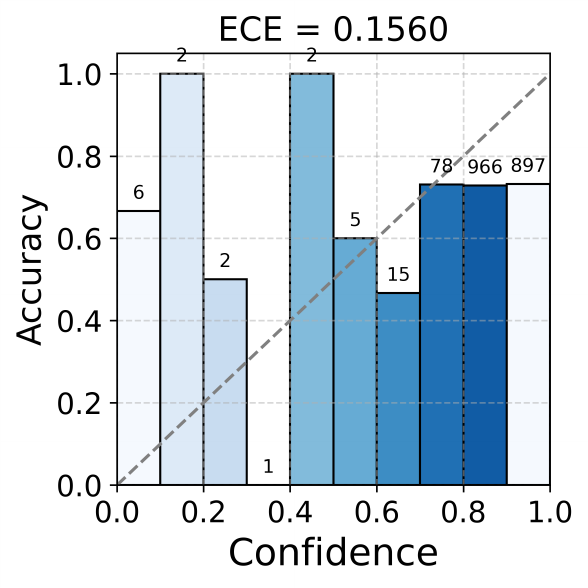}
  \caption{Self-Consistency}
\end{subfigure}
\par
\begin{subfigure}{0.23\textwidth}
  \centering
  \includegraphics[width=\linewidth]{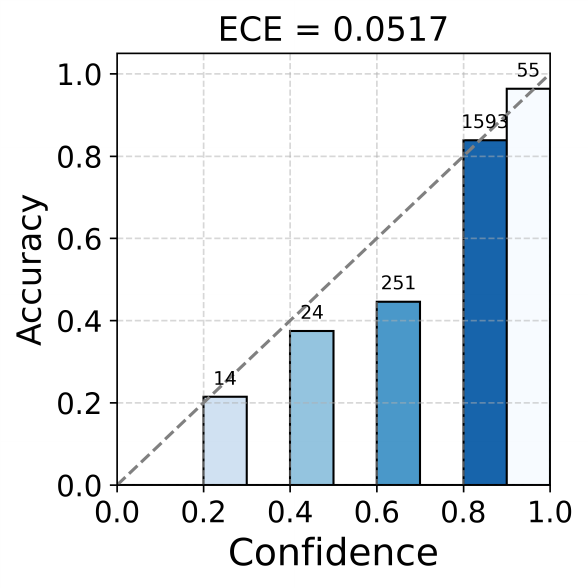}
  \caption{ConfTuner}
\end{subfigure}
\begin{subfigure}{0.23\textwidth}
  \centering
  \includegraphics[width=\linewidth]{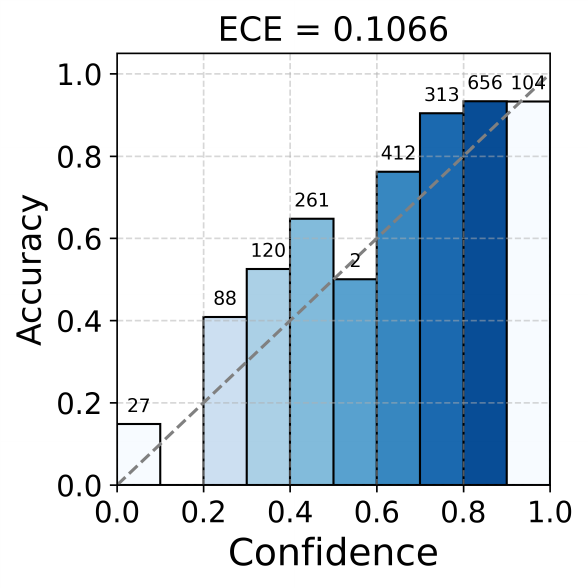}
  \caption{ADVICE}
\end{subfigure}
\begin{subfigure}{0.23\textwidth}
  \centering
  \includegraphics[width=\linewidth]{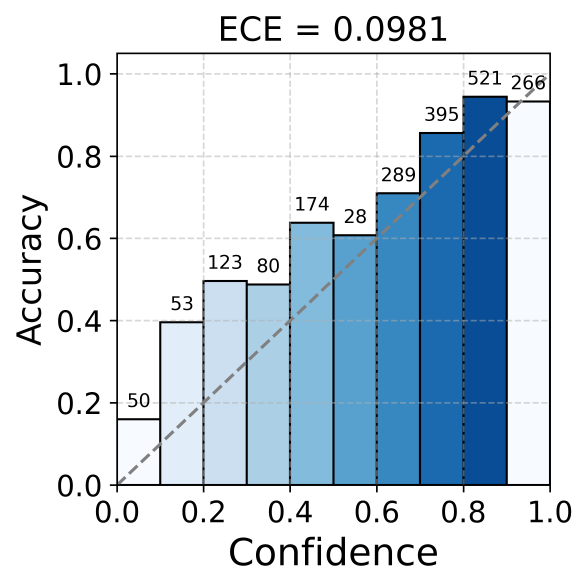}
    \caption{ADVICE \textit{w/} ConfTuner}
\end{subfigure}
\caption{Reliability diagrams of \textsc{Llama-3.1-8B-Instruct} on TriviaQA under the ScoreNumber setting, where numbers above each bin indicate the number of instances.}
\label{fig:reliability_diagram_appendix2}
\end{figure*}

\begin{table*}[b]
\small
\centering
\setlength{\tabcolsep}{4pt}
\begin{tabular}{ll}
\toprule
\textbf{Task} & \textbf{Prompt} \\
\midrule
Generation & \begin{tabular}[t]{@{}l@{}}
Provide only a short answer in the form of keywords to the following question.
\end{tabular} \\ 
\midrule
Multiple-Choice & \begin{tabular}[t]{@{}l@{}}
The following multiple-choice question has only one correct answer. Provide only \\
the option letter of the correct answer.
\end{tabular} \\
\bottomrule
\end{tabular}
\caption{These are task-dependent prefix prompts that are placed before the main prompt template.}
\label{tab:prefix_prompt}
\end{table*}

\begin{table*}
\small
\centering
\begin{tabular}{ll}
\toprule
\textbf{Type} & \textbf{Prompt} \\
\midrule
ScoreLetter   & \begin{tabular}[t]{@{}l@{}}
After your answer, provide a confidence score between A (very high confidence) \\
and E (very low confidence) which measures how confident you are in your answer. \\
Use the following format to respond:\\
``` \\
Answer: [Write your answer here.]\\
Confidence: [Write your confidence score here.]\\
```\\
If you cannot provide an answer, answer with `NO ANSWER`.
\end{tabular} \\ 
\midrule
ScoreNumber & \begin{tabular}[t]{@{}l@{}}
After your answer, provide a confidence score between 0 and 9 which measures how \\
confident you are in your answer, where 9 is the maximum. Never use 10. \\
Use the following format to respond:\\
``` \\
Answer: [Write your answer here.]\\
Confidence: [Write your confidence score here.]\\
```\\
If you cannot provide an answer, answer with `NO ANSWER`.
\end{tabular} \\
\midrule
ScoreText & \begin{tabular}[t]{@{}l@{}}
After your answer, provide one of the following confidence scores which measures \\
how confident you are in your answer: high, medium, low. \\
Use the following format to respond:\\
``` \\
Answer: [Write your answer here.]\\
Confidence: [Write your confidence score here.]\\
```\\
If you cannot provide an answer, answer with `NO ANSWER`.
\end{tabular} \\
\midrule
ScorePercent & \begin{tabular}[t]{@{}l@{}}
After your answer, provide a confidence score in percentage which measures how \\
confident you are in your answer. \\
Use the following format to respond:\\
``` \\
Answer: [Write your answer here.]\\
Confidence: [Write your confidence score here.]\\
```\\
If you cannot provide an answer, answer with `NO ANSWER`.
\end{tabular} \\
\midrule
ScoreFloat & \begin{tabular}[t]{@{}l@{}}
After your answer, provide a confidence score between 0.0 and 1.0  which measures \\
how confident you are in your answer. \\
Use the following format to respond:\\
``` \\
Answer: [Write your answer here.]\\
Confidence: [Write your confidence score here.]\\
```\\
If you cannot provide an answer, answer with `NO ANSWER`.
\end{tabular} \\
\bottomrule
\end{tabular}
\caption{Main prompt variations depending on verbalization type.}
\label{tab:main_prompts}
\end{table*}

\begin{table*}[t]
\small
\centering
\begin{tabular}{ll}
\toprule
\textbf{Type} & \textbf{Prompt} \\
\midrule
ConfTuner & \begin{tabular}[t]{@{}l@{}}
You will be asked trivia questions. Please respond to the best of your ability.\\
Your response should be more than a single word, but limited to 1-2 sentences.\\
Then please extract a single answer from the your response. \\
If no answer is present, please write "NONE". Finally, please provide your \\
confidence (0-9) to your answer.\\
\\
Here are some examples:\\
\\
Question: Who wrote Paradise Lost?\\
Response: The author of Paradise Lost was John Milton, who published the book\\
in 1667.\\
Final answer: John Milton\\
Confidence: 8\\
\\
Question: Which colonial power did Algeria gain independence from in 1962?\\
Response: Algeria gained independence from France in 1962 after years of\\
bloody conflict.\\
Final answer: France\\
Confidence: 9\\
\\
Question: How many planets are in our solar system?\\
Response: Please respond to the survey link below:\\
https://www.surveymonkey.com/r/5VZ7Z6P\\
Final answer: NONE\\
Confidence: 0\\
\\
Question: \{QUESTION\}\\
Response:
\end{tabular} \\
\bottomrule
\end{tabular}
\caption{Prompt for ConfTuner.
}
\label{tab:conftuner_prompt}
\end{table*}

\end{document}